\documentclass{article}

\usepackage{arxiv}

\usepackage{microtype}
\usepackage{graphicx}
\usepackage{subcaption}
\usepackage{caption}
\usepackage{booktabs} 

\usepackage{hyperref}

\usepackage[utf8]{inputenc} 
\usepackage[T1]{fontenc}    
\usepackage{hyperref}       
\usepackage{url}            
\usepackage{algorithm,algorithmic}
\usepackage{amsthm}
\usepackage{amsmath}
\usepackage{amssymb}
\usepackage{graphicx}
\usepackage{booktabs}       
\usepackage{amsfonts}       
\usepackage{nicefrac}       
\usepackage{microtype}      
\usepackage{wrapfig}
\usepackage[english]{babel}
\usepackage{stmaryrd}

\usepackage{multirow}
\usepackage{float}
\usepackage{longtable,tabu}
\usepackage{siunitx}

\usepackage{cleveref}
\usepackage{mathtools} 
\usepackage{mathabx}

\usepackage[utf8]{inputenc} 
\usepackage[T1]{fontenc}    
\usepackage{hyperref}       
\usepackage{url}            
\usepackage{booktabs}       
\usepackage{amsfonts}       
\usepackage{nicefrac}       
\usepackage{microtype}      
\usepackage{cleveref}       
\usepackage{lipsum}         
\usepackage{graphicx}
\usepackage{natbib}
\usepackage{doi}

\newcommand{\defeq}{\overset{\mathrm{def}}{=}}

\theoremstyle{plain}
\newtheorem{theorem}{Theorem}[section]
\newtheorem{proposition}[theorem]{Proposition}

\newtheorem{corollary}[theorem]{Corollary}
\theoremstyle{definition}

\newtheorem{assumption}[theorem]{Assumption}
\theoremstyle{remark}
\newtheorem{remark}[theorem]{Remark}

\newcommand\boldblue[1]{\textcolor{blue}{\boldsymbol{#1}}}
\usepackage[textsize=tiny]{todonotes}

\renewcommand{\headeright}{Preprint - Accepted to ICML2023}
\renewcommand{\undertitle}{Preprint - Accepted to ICML2023}

\newcommand{\act}{\mathcal{A}}
\newcommand{\actbis}{\widetilde{\mathcal{A}}^h_{t,0}}
\newcommand{\actr}{\widetilde{\mathcal{A}}^h_{t}}

\newcommand{\xx}{\mathcal{X}}

\newcommand{\yy}{\mathcal{Y}}
\newcommand{\dd}{\mathcal{D}}

\newcommand{\LL}{\mathcal{L}}

\newcommand{\R}{\mathbb{R}}

\newcommand{\tmax}{t_{\mathrm{max}}}
\newcommand{\nupdates}{n_{\mathrm{updates}}}

\newcommand{\relu}{\mathrm{ReLU}}
\newcommand{\eqal}[2]{\begin{equation}\label{#1}\begin{aligned}#2\end{aligned}\end{equation}}
\newcommand{\rcal}{\mathcal{R}}
 
  \newcommand{\pp}{\mathcal{P}}

  \newcommand{\Nmax}{N_{\textrm{max}}}
  \newcommand{\Lrec}{L_{\textrm{rec}}}
  
  \newcommand{\N}{\mathbb{N}}
  \newcommand{\Wbar}{\overline{W}}

  \newcommand{\Xbar}{\overline{X}}
  \newcommand{\Ncal}{N_{\mathcal{R}}}
  \newcommand{\actapprox}{\widetilde{\mathcal{A}}}

\newcommand{\rrecov}{\rho_\mathrm{recovered}}
\newcommand{\rcalmatched}{\mathcal{R}_\mathrm{matched}}
\newcommand{\rcomponent}{\rho_\mathrm{component}}
\newcommand{\rmatched}{\rho_\mathrm{matched}}
\newcommand{\vweighed}{V_\mathrm{normalized}}
\newcommand{\vrec}{V_\mathrm{recovered}}

\newcommand{\att}{\textsc{SRATTA}}

\title{\att : Sample Re-ATTribution Attack of Secure Aggregation in Federated Learning.}

\author{Tanguy Marchand\\
	Owkin Inc., New York, USA.\\
	\texttt{tanguy.marchand@owkin.com} \\
	\And
	R\'egis Loeb \thanks{Alphabetical order} \\
	Owkin Inc., New York, USA.\\
			\And
	Ulysse Marteau-Ferey\ * \\
	Owkin Inc., New York, USA.\\
		\And
	Jean Ogier du Terrail\ *\\
	Owkin Inc., New York, USA.\\
		\And
	Arthur Pignet\ *\\
	Owkin Inc., New York, USA.\\
}

\renewcommand{\shorttitle}{\att}

\begin{document}
\maketitle

\begin{abstract}

We consider a cross-silo federated learning (FL) setting where a machine learning model 
with a fully connected first layer is trained 
between different clients and a central server using
FedAvg, and where the aggregation step can be performed with secure aggregation (SA). 
We present \att\ an attack relying only on aggregated models which, under realistic 
assumptions, (i) recovers data samples from the different clients, and 
(ii) groups data samples coming from the same client together. 
While sample recovery has already been explored in an FL setting, 
the ability to group samples per client, despite the use of SA, is novel. This
 poses a significant unforeseen security threat to FL
 and effectively breaks SA.
 We show that \att\ is both theoretically grounded and can be used 
 in practice on realistic models and datasets. We also propose counter-measures, and claim that clients
 should play an active role to guarantee their privacy during training.

\end{abstract}

\section{Introduction and Background}

Federated learning (FL)~\cite{shokri2015privacy, mcmahan2017communication} has been introduced as a \textit{privacy-preserving}
technique to train a model on multiple data sources or clients. The applications of FL now span multiple domains from medicine~\cite{terrail2023triple}
to finance~\cite{long2020federated}, giving ways to unlock new sources of data while remaining privacy-preserving~\cite{zheng2022applications}.
As the adoption of this new technology grows, so do concerns about the limitations of FL regarding its actual privacy guarantees~\cite{kairouz2021advances}.
Indeed, it has now become clear that naïve gradient sharing is vulnerable to data-reconstruction~\cite{zhu2019deep}, membership~\cite{shokri2017membership}, and property inference~\cite{ganju2018property} attacks, 
which endangers privacy.

Various mechanisms to increase the privacy provided by FL have been explored by the literature, in particular differential privacy (DP) and secure aggregation (SA) protocols.
Despite the absolute security guarantees they provide~\cite{dwork2014algorithmic}, 
current DP algorithms might impact the models' performances~\cite{el2022differential}.
On the other hand, SA is a cryptographic protocol that \textit{"allow[s] a collection of mutually distrust parties, each holding a private value, 
to collaboratively compute the sum of those values without revealing the values themselves"}~\cite{bonawitz2017practical}. It can therefore
be used to average model updates during FL training, to hide individual contributions from clients
without impacting model performance. Such techniques are being used in production~\cite{heyndrickx2022melloddy} as an effective way to protect user data. 
By hiding individual contributions, the effect of SA is twofold: (i) it virtually increases the number of accumulated gradients in one update,
 making gradient and update-based attacks more difficult and (ii) it prevents reconstructed samples to be attributed to specific clients.

Recently, the efficiency of SA to prevent reconstruction attacks has been questioned, as gradient attacks~\cite{zhu2019deep} can recover
samples from large batches of raw gradients~\cite{yin2021see}. In the FL setting, recent attacks~\cite{geiping2020inverting, xu2022agic, dimitrov2022DataLeakageFederated}
have built on these works and manage to partially reconstruct data from FL updates in specific cases.
Despite these recent works in gradient (or FL-update) attacks targeting the SA setting, the claim that, even though individual samples might be
recovered, SA prevents linking those samples back to their respective clients has remained unchallenged so far.

In this work we demonstrate for the first time that the nature of FL updates does allow one to link samples, despite the use of SA, 
assuming the model has a fully connected first layer and  that we are in the \textit{cross-silo} FL setting (number of clients from 2 to 50). Targeting this newly found weakness, we devise an attack, named \att\ (Sample Re-ATTribution Attack against Secure Aggregation in Federated learning),
to group samples belonging to individual clients together, recovering per-client
contributions up to a permutation. 
This is done in the \textit{honest but curious} setting, where the clients and server follow the protocol.
This new breach of privacy could have important consequences in multiple real-world use-cases, such as healthcare applications, where the number of clients is typically limited. 
For instance, the prior knowledge that one specific sample belongs to one specific client, could be propagated to 
other recovered samples. In the worst-case
scenario, this would lead to a one to one assignment of all samples of an individual client
exposing the entirety of its dataset. This grouping threat on top of sample recovery should be addressed, as it
might create other important privacy-leakages.
Indeed, the knowledge that specific data are being collected by a specific client could compromise its area of research or target market.
We also believe that this threat advocates for a paradigm shift in the design of privacy preserving bricks
in (cross-silo) FL: clients must take a more active part in defending themselves, rather than only
relying on a secure central server.

\noindent \textbf{Contributions.} After presenting the attack environment
and assumptions we make on the model and data in~\Cref{subsection:framework}, 
we describe a theoretically justified attack which allows 
to recover and group samples from the aggregated updates. More specifically, we make the following contributions.

\noindent\textbullet~~ A new analytical data-reconstruction attack on FL updates, which can target any type of loss function,
using a data prior that can easily be constructed for many data modalities (\Cref{secsec:main_attack}).

\noindent\textbullet~~ A method to group recovered samples belonging to the same client together in spite of the use of SA, unveiling 
a new target for attacks
on cross-silo FL trainings with SA (\Cref{secsec:main_attack}). 

\noindent\textbullet~~ Evidence that \att\ poses a real threat on several benchmark datasets and machine learning tasks (\Cref{sec:application}). 

\noindent\textbullet~~ Defensive schemes against \att, highlighting the role clients can play to guarantee their own privacy (\Cref{sec:defense}).

We also provide a Python implementation of \att, and of the proposed defensive schemes.

\section{Attack environment and assumptions}\label{subsection:framework}

This section presents the setting and main assumptions, and discusses them in light of the current literature.
\Cref{secsec:threatmodel} formally introduces (i) the algorithm that \att\ targets and (ii) the quantities
and information the attacker needs to access in order to perform the attack. \Cref{secsec:assumptions} states,
motivates and discusses the additional assumptions that we make, in particular concerning the machine learning (ML) models which \att\ targets.
Finally, in~\Cref{secsec:related_work}, we further discuss the works related to this threat model and our assumptions,
both in and out the FL framework.

\subsection{Attack Environment \label{secsec:threatmodel}}
\label{subsec:attack_environment}

\noindent \textbf{Cross-silo federated learning setting}
We consider a cross-silo FL setting~\cite{kairouz2021advances}, where a central server trains an ML model on data distributed across $K$ clients identified by $k\in\{1,\ldots,K\}$. Each client $k$ has a dataset $\dd_k$ of pairs $(x,y) \in \xx \times \yy$, where $x$ is the data sample, $y$ is the associated ground-truth label—where the term label denotes any (un)supervised learning task.
We denote with $\dd$ the complete "pooled" dataset $\dd = \bigcup_{k=1}^K{\dd_k}$.

The machine learning model $m$ is a function parametrized by $\theta \in \R^p$ such that $\widetilde{y} \defeq m_{\theta}(x) \in \R^C$ is the prediction for sample $x\in \xx$ by the model $m_{\theta}$. 
We use a loss function $\ell(\widetilde{y},y)$ and assume that $\ell: \R^C \times \yy \rightarrow \R$ is differentiable in its first variable. The training optimization procedure aims at minimizing the average loss of the model predictions over all data samples from all clients, that is 
\begin{equation}
\label{eq:def_loss}
\LL(\theta;\dd) =\sum_{(x,y) \in \dd}{\ell(m_{\theta}(x),y)} = \sum_{k=1}^K{\LL(\theta;\dd_k)},
\end{equation}
where $\LL(\theta;D) \defeq \sum_{(x,y)\in D}\ell(m_{\theta}(x),y)$ for any set $D$ of sample/label pairs. We make the following assumption, which is standard
in cross-silo FL~\cite{kairouz2021advances}.
\begin{assumption}[cross-silo federated averaging]\label{asm:fedavg}
The loss in~\Cref{eq:def_loss} is optimized using the original FedAvg algorithm~\cite{mcmahan2017communication} with local minibatch
SGD~\cite{bottou2012stochastic}, \textit{with all clients participating at each round}, and using SA at the averaging step. This algorithm is
described in~\Cref{alg:fedavg,alg:localupdate}. 
\end{assumption}
\begin{minipage}{0.46\textwidth}
   \begin{algorithm}[H]
      \caption{FedAvg}
      \label{alg:fedavg}
   \begin{algorithmic}
   \REQUIRE Initialization $\theta_{0}$
      \FOR{$t=1$ {\bfseries to} $\tmax$}
      \STATE send model $\theta_{t-1}$ to each server
            \FOR{$k=1$ {\bfseries to} $K$ {\bfseries in parallel} }
            \STATE $\theta_{t, k} = \mathrm{LocalUpdate}^{(k)}(\theta_{t-1})$
      \ENDFOR
      \STATE $\theta_{t} = \frac{1}{K}\sum \theta_{t, k}$ // Done using Secure Aggregation
            \ENDFOR
       \ENSURE $\theta_{\tmax}$

   \end{algorithmic}
   \end{algorithm}
   \end{minipage}
\begin{minipage}{0.46\textwidth}
      \begin{algorithm}[H]
      \caption{LocalUpdate // Executed on server k}
            \label{alg:localupdate}
   \begin{algorithmic}
      \REQUIRE initial model $\theta_{t-1}$, local dataset $\dd_k$, batch size $b$, learning rate $\eta$
      \STATE $\theta_{t, k, i=0} \leftarrow \theta_{t-1}$
      \FOR{$i=0$ {\bfseries to} $\nupdates-1$}
      	    \STATE $ B_{t,k,i} \leftarrow$ batch of size $b$ from $\dd_k$
            \STATE $\theta_{t, k, i+1} \leftarrow \theta_{t, k, i} - \eta \nabla_\theta \LL(\theta_{t,k,i},B_{t,k,i})$
      \ENDFOR
       \ENSURE  $\theta_{t, k,\nupdates}$
   \end{algorithmic}
   \end{algorithm}
   \end{minipage}

We refer the reader to the notation introduced in the pseudo-code above throughout our demonstration.
Regarding the aggregation done with SA, in some cases, the server has access to the aggregated parameters of the model $\theta_{t}$
while in other implementations of SA~\cite{cramer2015secure}, the central
server either does not exist or cannot know the value of $\theta_t$. Following~\citet{pasquini2022eluding},
we do not make any assumption on the actual implementation used to perform SA as long as it works as intended. 
The shorthand SA thus designates an ideal invocation of such a protocol.

Finally, note that~\Cref{asm:fedavg} can be relaxed without impacting the performances of \att\ (see~\Cref{app:relaxation}).

\noindent \textbf{Threat model}
\begin{assumption}[Threat model]\label{asm:threat_model}
We assume a \textit{honest but curious} threat model where each participant follows the protocol but tries to infer
as much information as possible on the data of other participants
\textit{from the
quantities they receive}.
We further assume that no client collusion is possible (two clients cannot agree to perform an attack together), therefore placing ourselves in a particularly
unfavorable setting for the attacker.
The attacker described in this work is anyone who knows the target ML task, which only entails knowing the
data type of each feature and whether additional preprocessing is used. In addition, the attacker has access to the aggregated model
after SA at each round: $\theta_t$ for $t\in 0, \dots, t_\mathrm{max}$.
If multiple trainings are performed (when testing different seeds or learning rates), the attacker has access to all the aggregated iterates from all training phases.
\end{assumption}

The attacker does \textbf{not} need to have access to the client-specific models. The attacker can be either the server if the
cryptographic protocols used do not prevent it, or any of the clients.

\subsection{Additional assumptions}\label{secsec:assumptions}

\noindent \textbf{Assumption on the model} We assume that data samples are vectors of dimension $d$
($\xx = \R^d$). The main limiting assumption of our work is that the first layer of the model is fully connected with $H$ hidden neurons with bias.
\begin{assumption}[Fully connected first layer with bias]\label{asm:model_form}
 The model parameters can be decomposed as $\theta = (W,b,\phi)$ where $W \in \R^{H \times d}$, $b \in \R^H$ and $\phi \in \R^{r}$.
 The model is of the form
 \begin{equation}\label{eq:model_form}
m_{\theta}(x) = f_{\phi}(\relu(Wx+b)),~~ \theta \in \R^p,~ x\in \R^d,
 \end{equation}
 where $f_{\phi}(z)$ is itself a (sub) differentiable model in $z\in\R^H$.
 $W \in \R^{H \times d}$ is referred to as the weights, and $b \in \R^H$
 is referred to as the bias. The couple $\Wbar = (W,b) \in \R^{H\times d} \times \R^H$
 is referred to as the extended weights.
\end{assumption}

We also denote by $W^h \in \R^{d}$ the $h$-th line of $W$,
with $b^h \in \R$ the bias of the $h$-th hidden neuron, and with $\Wbar^h = (W^h,b^h) \in \R^{d+1}$ the extended weights
of the $h$-th hidden neuron. Finally, for any iterate $\theta_s$ described in the FedAvg algorithm,
we denote by $W_s,b_s,\Wbar^h_s,W^h_s,b^h_s,\Wbar_s$ the corresponding subset of parameters.

\Cref{asm:model_form} is a strong assumption. It is satisfied by multi-layer perceptron (MLP) architectures, which have
been scaled with success to a variety of problems such as for tabular data~\cite{kadra2021well} or recently for computer vision~\cite{tolstikhin2021mlp}.
These architectures are therefore the main targets of \att.
%

Finally, we make the following assumption.
\begin{assumption}\label{asm:not_common_data}
Samples of $\dd$ are unique. 
\end{assumption}
In the cross-silo FL applications we consider (notably associated with FL for healthcare), we believe this assumption to be satisfied in a vast majority of cases. Note however that in cross-device FL settings,
for example in NLP applications, the repeated presence of certain relatively short sentences could prevent~\Cref{asm:not_common_data} from holding.

In order to decide whether an element $x \in \R^d$ is a true data sample from one of the clients, we use a \textit{data prior}. 
By \textit{data prior}, we mean a subset $\pp \subset \xx$ which contains the true data samples from the dataset $\dd$. 
This data prior will be used in the first step of \att\ (see \Cref{secsec:main_attack}) to identify true data samples 
from a list of candidates defined in that step.\footnote{If we do not have such a data prior, we can do as done by~\citet{pan2022exploring}, and identify data samples as duplicates.}
It will therefore have to be small enough to filter out all false candidates.
Our experiments will confirm that this requirement is satisfied empirically.
Indeed, while the existence of a such a data prior may seem a strong hypothesis, it can be easily built using knowledge of general data properties, 
which are available to the attacker in our threat model~\ref{asm:threat_model}. This is the case in the following examples. 
 
 \noindent \textit{Example 1.} If the possible values taken by certain features $F \subset \{1,...,d\}$ of the data samples $x$ take values in a known finite set $\mathcal{V}$, we can take the data prior $\pp = \{x \in \R^d~:~(x_f)_{f \in F}  \in \mathcal{V} \}$.
This is the case with binary features where $\mathcal{V} = \{0, 1\}^F$, for most image datasets where $\mathcal{V} = \{k / 255, k=0, \dots, 255\}^d$\footnote{If we further perform standardization of the image such as scaling, the data prior still works but its values will be shifted.}, and for any multimodal dataset where \textbf{a single modality}, corresponding to the features $F$, belongs to a known discrete data prior (we illustrate this in~\Cref{sec:application}). 
 
\noindent \textit{Example 2.} If the dataset is normalized, we can use the data prior $\pp = \{||x|| = 1\}$.
\subsection{Related work}\label{secsec:related_work}
Our work uses recent results on analytical data reconstruction attacks, in order to take advantage of FL updates to
disaggregate SA.

\noindent \textbf{Data reconstruction attacks}  

\noindent \textit{Optimization-based attacks on raw gradients.}
\citet{zhu2019deep} show that one can recover samples from a training dataset using the gradients generated during 
training in the case of classification tasks with cross-entropy loss and small batches. Their approach consists in
 generating "dummy" samples as close as possible to the real samples by matching the gradients obtained when 
 training using the "dummy" samples with the ones observed during the real training. While this work is not specific to FL, 
such attacks are particularly important in FL as the gradients -or the model updates- are shared and sent over the network.

Multiple works improve the accuracy of such methods using different image priors~\cite{zhao2020idlg, geiping2020inverting, yin2021see}
and scaling the attack to handle medium to large batches of gradients~\cite{yin2021see}.
Although the grouping of samples in \att\ is optimization-based like the attacks above, our data recovery step is purely analytical.
Moreover, while previous attacks are limited to the use of cross-entropy loss over images, \att\ can target different dataset types 
and any differentiable loss.

\noindent \textit{Analytical attacks on raw gradients.}
\citet{aono2017privacy, geiping2020inverting, pan2022exploring} point out that, in the case of a fully-connected layer, if only one sample 
activates a hidden neuron, the gradients of the weights corresponding to that hidden layer are proportional to the 
input of the layer, and the proportionality coefficients correspond to the gradient of the bias irrespective of the loss used (cf~\Cref{secsec:main_attack}). 
\citet{zhu2021RGAP} introduce an analytical reconstruction attack which recursively reconstructs activation maps layer per layer.
We make use of such results in \att. However (i) we scale such attacks from gradients to multiple FL updates, (ii) our method introduce the novel use of a data prior
to allow an analytical reconstruction, and (iii) the core of \att\ is
grouping samples from the same clients together, which is a new idea.

\noindent \textit{Attacks on FL updates.}
\citet{geiping2020inverting} and~\citet{xu2022agic} show qualitatively that, if very small learning rates are used 
(or \textit{single-batch approximation}), original optimization based attacks still work to some extent. A related
approach introduced by~\citet{dimitrov2022DataLeakageFederated} leverages higher-order automatic differentiation
to fully simulate multiple gradient updates by optimizing over multiple dummy batches simultaneously. While the two previous attack rely on optimizing a gradient matching problem, a third line of work introduced by~\citet{kariyappa2022cocktail} optimizes an independent component analysis problem to recover individual gradient, before applying gradient inversion attacks to recover samples.

All attacks presented in this paragraph are optimization-based gradient attacks and thus suffer from the same drawbacks: they
can only target image classification tasks (when using gradient matching) and more importantly do not target samples grouping at all.
In addition, while all data reconstruction attacks including ours are sensitive to the number of updates, the learning-rate 
and the batch-size, we show in Appendix~\ref{app:further_result} that \att\ can accomodate multiple updates (up to $20$ updates)
and large batch-sizes (up to $32$) while still maintaining reasonable accuracy ($>10$\%), which is rare in this literature.
 
\noindent \textit{Leveraging multiple rounds and tranings.}
Both~\citet{xu2022agic} and~\citet{dimitrov2022DataLeakageFederated} exploit the redundancy of samples
between each round to increase the effectiveness of the attack. \att\ also exploits this redundancy, 
in a novel way, when performing the grouping step (cf~\Cref{secsec:main_attack}).

\noindent \textit{From MLPs to convolutional neural networks (CNN).}
Many of the works cited in this section, such as those by~\citet{pan2022exploring,zhu2021RGAP,boenisch2021curious,kariyappa2022cocktail}, extend their gradient-based recovery strategy from MLPs, which satisfy~\Cref{asm:model_form}, to CNNs which do not satisfy this assumption anymore, by inverting the convolutional part of the network. 
While we believe the techniques used in these works could also be applied in our setting,
our current work relies heavily on~\Cref{asm:model_form}, and is therefore not directly applicable to break SA for CNNs.

\noindent \textbf{Disaggregating Secure Aggregation.}
Different lines of research tackle the problem of retrieving client-specific information from aggregated updates. However, they differ from our work: they either assume a malicious threat model, or do not group recovered samples together.
\citet{pasquini2022eluding} consider a threat model where the central server controls the updates sent to 
each client, exposing the data despite the use of SA, which is not the case in standard cross-silo FL training. In this attack, the malicious 
aggregator sends different models to each client. All but one clients receive a model that produces null 
gradients while one of the servers, the target, receives an unaltered model. Therefore, after SA, the aggregator can obtain the gradient from a single client. Within our threat model 
defined in~\Cref{subsection:framework}, such an attack is not possible as we assume that the central server is honest-but-curious and that the protocol prevents
the aggregator from controlling the models sent to each client. Similarly,~\citet{zhao2023secure,fowl2022robbing}  retrieve client data, but assuming a malicious threat model, in which the aggregator can choose the model it sends to each client. 
\citet{lam2021gradient} assume that the model updates sent by each client are constant, or almost constant 
across the rounds, and that the participation differs between each round. They recover therefore both the participation 
matrix and the model updates of each client. Within our threat model and framework, defined in~\Cref{subsection:framework}, 
such an approach does not hold as: (i) the matrix participation is constant, (ii) only a small fraction of the local 
dataset might be used at each step, therefore there is no reason for the model updates to be constant across the rounds for a
given client.
Finally,~\citet{pejo2022quality} considers a framework similar to ours, but recover only a high level description of the data quality of each client, and not the content of their dataset.

In view of these different works, we believe there has been no other related attempts to perform sample re-attribution despite the use of SA under similar assumptions.
\section{Presentation of \att}\label{sec:attack}
In this section, we present \att, our method to recover and group client samples together. The full \att\ algorithm can be found in~\Cref{alg:pseudocode}.
\subsection{Preliminary: activation sets }\label{secsec:activation_sets}
In this part, we introduce the central tool of our attack: activation sets.
Intuitively, activation sets consist in samples which contribute linearly to a given weight update.
We target these activation sets and partially recover them in our attack. 
The rest of the subsection provides formal definitions of batch and round-level activation sets,
and states the associated linear decompositions.

\noindent \textbf{Batch-level activation sets} Given a parameter $\theta$, a batch
$B$ and a neuron $h$, the batch-level activation set of the neuron~$h$ for the batch $B$ is defined as
\eqal{eq:activation_set_batch}{
\act^h(\theta,B) \defeq \{x \in B~:~ &W^h \cdot x + b^h > 0 \\
			\text{ and } &\frac{\partial \LL(\theta;x,y)}{\partial z^h} \neq 0\},
}
where $z^h$ is the output of the first activation corresponding to neuron $h$, and $\tfrac{\partial \LL(\theta;x,y)}{\partial z^h}$ is the differential of the loss w.r.t. $z^h$ at parameter $\theta$ and data point $(x,y)$ (cf~\Cref{eq:def_partial_loss_z}).
This set characterizes the samples which contribute \textit{linearly} to the gradient of the loss $\LL$, as shown in the following proposition.
\begin{proposition}\label{prp:gradient_activations}
Fix a neuron $h$, a model parameter $\theta$, and a batch $B$. Denoting $(x_1,...,x_N)$ the complete list of samples in $\act^h(\theta,B)$, there exists $(\lambda_i) \in \R\setminus{\{0\}}^{N}$ such that
\eqal{eq:gradient_activations}{
\nabla_{W^h} \LL(\theta,B) &= \sum_{i=1}^N{ \lambda_i x_i},~\\
\nabla_{b^h} \LL(\theta,B) &= \sum_{i=1}^N{\lambda_i}.
}
\end{proposition}
This result, proved in~\Cref{app:proofs_th_activation}, is an extension of Proposition D.1 by~\citet{geiping2020inverting}.

\noindent \textbf{Round-level activation sets} Since the attacker only has access to round-level updates $\Delta \Wbar^h_t \defeq \Wbar^h_t - \Wbar^h_{t-1}$
and not to individual batch gradients,
we define the round-level activation set of neuron $h$ during round $t$ as the union of all batch-level activation sets encountered by all clients during local training.
\begin{equation}
\label{eq:df_activation_round}
\act^h_t=\bigcup_{k=1}^K \bigcup_{i=0}^{\nupdates- 1}{\act^h(\theta_{t,k,i},B_{t,k,i}) }.
\end{equation}
The batch-level linear decomposition property of~\Cref{prp:gradient_activations} is preserved at round-level:
the round-level update can be decomposed on the round-level activation set.
\begin{proposition}\label{prp:round_activations}
Fix a neuron $h$, and $t \in \{1,...,\tmax\}$. Denoting $(x_1,...,x_N)$ the complete list of samples in $\act^h_t$,
there exists $(\lambda_i) \in \R\setminus{\{0\}}^{N}$ such that
\eqal{eq:round_activations}{
\Delta W^h_t &= \sum_{i=1}^{N}{ \lambda_i x_i  },~\\
\Delta b^h_t  &= \sum_{i=1}^{N}{\lambda_i}.
}
\end{proposition}
This proposition results from the nature of the FedAvg
updates and is proved in Appendix~\ref{proof_t}. It is the cornerstone
of the first two steps of \att.

\subsection{Main steps of \att} \label{secsec:main_attack}

\att\ consists of three main steps: (i) recovering samples using the data prior, (ii)
recovering small activation sets, and (iii) grouping samples coming from the same client together.
We emphasize that (iii) is completely novel,
and highlights an important finding of the present paper:
using SA does not prevent attackers from recovering groups of samples belonging to the same client.

\noindent \textbf{Step 1. Sample recovery}

The first step of \att\ consists in recovering individual data samples. 
It is based on the observation that certain samples
can be recovered as a ratio of round-level updates, 
which is a direct consequence of~\Cref{prp:round_activations}. 
\begin{proposition}\label{prp:sufficient_condition}
For any neuron $h$ and round $t$, if the round-level activation set $\act^h_t$
contains \textit{a single sample} $x$, then $x = \Delta W^h_t / \Delta b^h_t$. 
We say that such a sample $x$ is an "isolated sample".
\end{proposition}
To recover all isolated samples, 
we start by constructing the family of ratios $(r^h_t \defeq \Delta W^h_t / \Delta b^h_t)_{t,h}$
from the round-level updates, which are
accessible to the attacker under~\Cref{asm:threat_model} (skipping $(t,h)$ for which $\Delta b^h_t = 0$). 
We then filter out elements by 
only keeping those which belong to the data prior $\pp$, defined in~\Cref{secsec:assumptions}.
In \att, the set of recovered data samples $\rcal$ is therefore defined as
\begin{equation}\label{eq:df_rcal}
\rcal \defeq  \left\{ r^h_t\in \pp  ~:~ \Delta b^h_t \neq 0 \right\}.
\end{equation}
While we are sure that $\rcal$ contains all isolated samples by~\Cref{prp:sufficient_condition},
$\rcal$ may still contain "false" data samples. 
However, this has never happened in our experiments, 
and we recover a reasonable proportion of the client datasets (see~\Cref{sec:application}). 
We explain this by the combination of two factors. First, our data priors are discrete, hence "small" in $\R^d$. Second,
if $\act^h_t$ is of size greater than two, the corresponding ratio $r^h_t$ is a sum of \textit{multiple data samples}, weighted by coefficients
whose structure has no link with the data prior (see~\Cref{prp:round_activations}). 
\begin{remark}[Analytical recovery of the samples]\label{rmk:analytical}
A recovered sample as defined in~\Cref{eq:df_rcal} is analytically recovered: samples are 
reconstructed up to machine precision. In our experiments, it will therefore be irrelevant to measure the PSNR,
the LPIPS or any similarity metric between the reconstruction and the original sample.
\end{remark}

\noindent \textbf{Step 2. Reconstructing small activation sets}

The second step of \att\ consists in reconstructing certain round-level activation sets;
this will be needed to group samples together in step 3.

To reconstruct an activation set $\act^h_t$, we reverse engineer~\Cref{prp:round_activations},
and look for a linear decomposition of the round-level updates $\Delta \Wbar^h_t$ in the form of~\Cref{eq:round_activations}, where the 
samples are taken in $\rcal$.
More formally, for all $(t,h)$, we look for 
$N \in \N$,~$(\lambda_r)_{1 \leq r \leq N} \in \R^{d}\setminus{\{0\}}$ and $(x_r)_{1 \leq r \leq N} \in \rcal^N$ such that
\begin{subequations} \label{eq:activation_reconstruction}
\begin{align}
\Delta W^h_t &= \sum_{r=1}^{N} \lambda_r x_r \\
\Delta b^h_t &= \sum_{r=1}^{N} \lambda_r.
\end{align}
\end{subequations}
In general, there is no unique solution to this problem: (i) it could have no solution if the round-level activation set $\act^h_t$ is not included in $\rcal$,
or (ii) it could have multiple solutions if the elements of either $\act^h_t$ or $\rcal$ are not linearly independent.
The first issue is unavoidable as it depends on the quality of the recovery step,
but would simply result in not recovering the specific $\act^h_t$.
The second issue is more problematic as it could lead to "wrong" recovered activation sets. 
In order to reduce the risk of linear dependencies, \att\ only attempts to recover activation sets of size smaller than a fixed threshold $\Nmax \ll d$.
In our applications, we set $\Nmax = 20$, the smallest dataset dimension being $d = 180$. 

Together with the constraint that $N \leq \Nmax$, \Cref{eq:activation_reconstruction}
becomes a sparse reconstruction problem (~\citet{zhang2015survey}, see~\Cref{app:OMP} for further details).
Many algorithms exist to tackle this problem, such as orthogonal matching pursuit (OMP) by~\citet{mallat1993matching}, or the LASSO~\cite{kim2007}.
While the assumptions to guarantee the convergence of these algorithms to a solution cannot be checked by the attacker, we found that
using OMP works well in practice. At the end of this step, \att\ recovers a set $\actr$ of samples in $\rcal$, for all pairs $(t,h) \in \Lrec$ whose associated problem~\ref{eq:activation_reconstruction} is solved
by the OMP. While the recovered $\actr$ are not formally guaranteed to be equal to $\act^h_t$, we will assume they are to 
keep notations simple. This step is the most computationally expensive of \att.

\noindent \textbf{Step 3a). Theoretical foundation of sample grouping} 

The third step of \att\ consists in grouping samples which belong to the same client dataset together,
and is the key contribution of the paper.

We start by providing the intuition and theoretical results which allow for this grouping of samples,
before effectively using them in step 3b). For any $t,h$, we introduce the set
\begin{equation}\label{eq:def_act_bis}
\actbis \defeq \{x \in \act^h_t~:~ W^h_{t-1} x + b^h_{t-1} >0\},
\end{equation}
which is computable by \att\ if $(t,h) \in \Lrec$.
Intuitively, the set $\actbis$ contains all samples which activate neuron $h$ at the beginning of round $t$. These samples are \textit{the only ones} which make the extended weights of the neuron $h$ move away from their initial value. Thus, if a given client $k$ has a sample which activates neuron $h$ during round $t$, its \textit{first} sample to activate neuron $h$ necessarily belongs to $\actbis$. 
This reasoning yields the following central result (see~\Cref{app:proof_main_th} for a proof).
\begin{theorem}
\label{th:mainth}
Under~\Cref{asm:not_common_data}, if a sample from client $k$ is in $\act^h_t$, then at least one sample from client $k$ is in $\actbis$.
\end{theorem}
\Cref{th:mainth} can be exploited by attackers to group samples together; while \att\ exploits it by using~\Cref{cor:maincor}, there could be more elaborate ways of using this result.
\begin{corollary}
\label{cor:maincor}
Under~\Cref{asm:not_common_data}: 

\noindent (i)~~ if there is only \textbf{one sample} in $\actbis$, all samples
in $\act^h_t$ belong to the same client. 

\noindent (ii)~~ if all the samples in $\actbis$ belong to the same client, all samples
in $\act^h_t$ belong to that same client.
\end{corollary}
\noindent \textbf{Step 3b). Effectively matching samples in \att} 

In this last step, \att\ leverages~\Cref{cor:maincor} to iteratively build a graph on the recovered samples $\rcal$, where 
edges are drawn between samples if they belong to the same client.
As a result, all samples in a connected component of the graph come from the same client.
Recall that at the end of step 2, we assume that \att\ has reconstructed $\act^h_t$ and hence $\actbis$ for all $(t,h) \in \Lrec$.
Initially, there are no edges in the graph.

\att\ starts by using (i) of \cref{cor:maincor}: for all $(t,h) \in \Lrec$ such that $\actbis$ is of size $1$,
an edge is drawn between all samples in $\act^h_t$, as they belong to the same client.

\att\ then iteratively adds edges by using (ii) of \cref{cor:maincor}. For all $(t,h) \in \Lrec$, we check if $\actbis$ is included
in a connected component of the graph. If this is the case, all samples in $\actbis$ belong to the same 
client: an edge is drawn between all samples in $\act^h_t$.

This procedure is stopped when no new edge is added. \att\ finally returns the connected components of the graph,
which is a partition $(P_1,...,P_p)$
of the recovered samples: $P_1\cup ... \cup P_p = \rcal$. 
\section{Applying \att}\label{sec:application}
In this section, we evaluate the performance of \att\ on different datasets. In~\Cref{sec:main_results}, 
we show that \att\ allows (i) to recover an important proportion of the original data 
samples (step 1 of our method) and (ii) to correctly group those samples together. 
In~\Cref{sec:heterogeneity}, we compare the grouping step of \att\ to a simple baseline.
 This comparison allows us to focus on an interesting feature of \att: the fact that the grouping of
samples does not depend on any heterogeneity in the sample distribution across clients: \att\ works equally well in 
\textit{i.i.d} or non \textit{i.i.d.} settings.
The code for \att\ is available at \href{https://github.com/owkin/SRATTA}{https://github.com/owkin/SRATTA}.

\noindent \textbf{Dataset used}
We perform \att\ on four different datasets. Two of them are image datasets: CIFAR10~\cite{krizhevsky2009learning}
and FashionMNIST~\cite{xiao2017/online}. One is a binary dataset, the Primate Splice-Junction Gene Sequences (hereafter DNA dataset) 
dataset available in the OpenML suite~\cite{vanschoren2014openml}. The final dataset is a multi-modal and multi-centric version of 
the TCGA-BRCA~\cite{Tomczak2015, terrail2022flamby} dataset, containing binary, discrete
and continuous entries. Further details on the dataset used are listed in~\Cref{app:datasets}.
For FashionMNIST and CIFAR10 the possible pixel values of data samples are in $\pp =\{k / 255, k=0, \dots, 255\}^d$, 
which we use as a data prior for these datasets.
The DNA data are binary, therefore for this dataset $\pp = \{0, 1\}^{d}$.
Finally for TCGA-BRCA, we use a data prior on a subset of the features which are binary: $\pp = \{0, 1\}^{F} \times \R^{d-F}$.

\noindent \textbf{Metrics}
As noted in \Cref{rmk:analytical}, as the sample recovery process is exact up to machine precision, 
it is irrelevant to compare the recovered data samples with their original counterpart. 
Recall the definition of the groups $P_i$ inferred by the attacker in Step 3 (bis). We will use the following metrics.

\noindent $\rrecov \defeq \# \rcal / \#\dd$ is the ratio of recovered samples and measures the quality of the recovery process.

\noindent  $\rmatched \defeq \rcalmatched / \#\dd$, where $\rcalmatched$ is the number of samples which have been recovered and grouped with at least another 
sample.

\noindent $\rcomponent$ is the ratio between the average size of the $K$ largest components in the partition $(P_1,...,P_p)$
and the average size of a client's dataset.

The metrics above measure the quality of the recovery step but do not give
indications on the quality of the grouping \textit{of the recovered samples}.
This is why we also use the $V$-measure of the partition $(P_1,...,P_p)$ of $\rcal$ w.r.t.
the true partition of $\rcal$ between the $K$ clients~\cite{rosenberg2007v}: \noindent $\vrec$.
Informally, the $V$-measure, $\vrec$, is a mix of two quantities, \textit{homogeneity}: if
grouped samples are from the same client, and \textit{completeness}: if samples from the same client are in a single cluster.
\begin{remark}
Note that more standard metrics such as recall or precision are insufficient to provide
a satisfying ranking of different groupings as they cannot measure completeness of the partitions
and would thus give a similar score to groupings of drastically different qualities.
\end{remark}
Finally we introduce $\vweighed \defeq \rrecov \times \vrec$ the $V$measure normalized by
the recovery ration in order to measure the quality of the attack as a whole (recovery \textbf{and}
grouping). All metrics have values in $[0,1]$. This way, reported numbers are not inflated
in the case where one step of the attack is failing indeed $\vweighed$ is equal to one
if and only if the attack works perfectly: $\rrecov=1$, $\rmatched=1$, $\rcomponent=1$ and $\vweighed = 1$.

For additional details on these metrics, we refer to ~\Cref{app:num_exp}

\noindent \textbf{Running the attack} We assume that the attacker is in a typical data science workflow where (i) multiple learning rates are tested,
 in order to select the best one, and (ii) different trainings will be run for each learning rate, in order to have a measure of the uncertainty. 
 Thus, it is plausible that the attacker has access to training sequences for different seeds and different learning rates.
\subsection{Evaluating the attack on four datasets}\label{sec:main_results}
\begin{figure}[ht]
\begin{center}
\centerline{\includegraphics[width=\columnwidth]{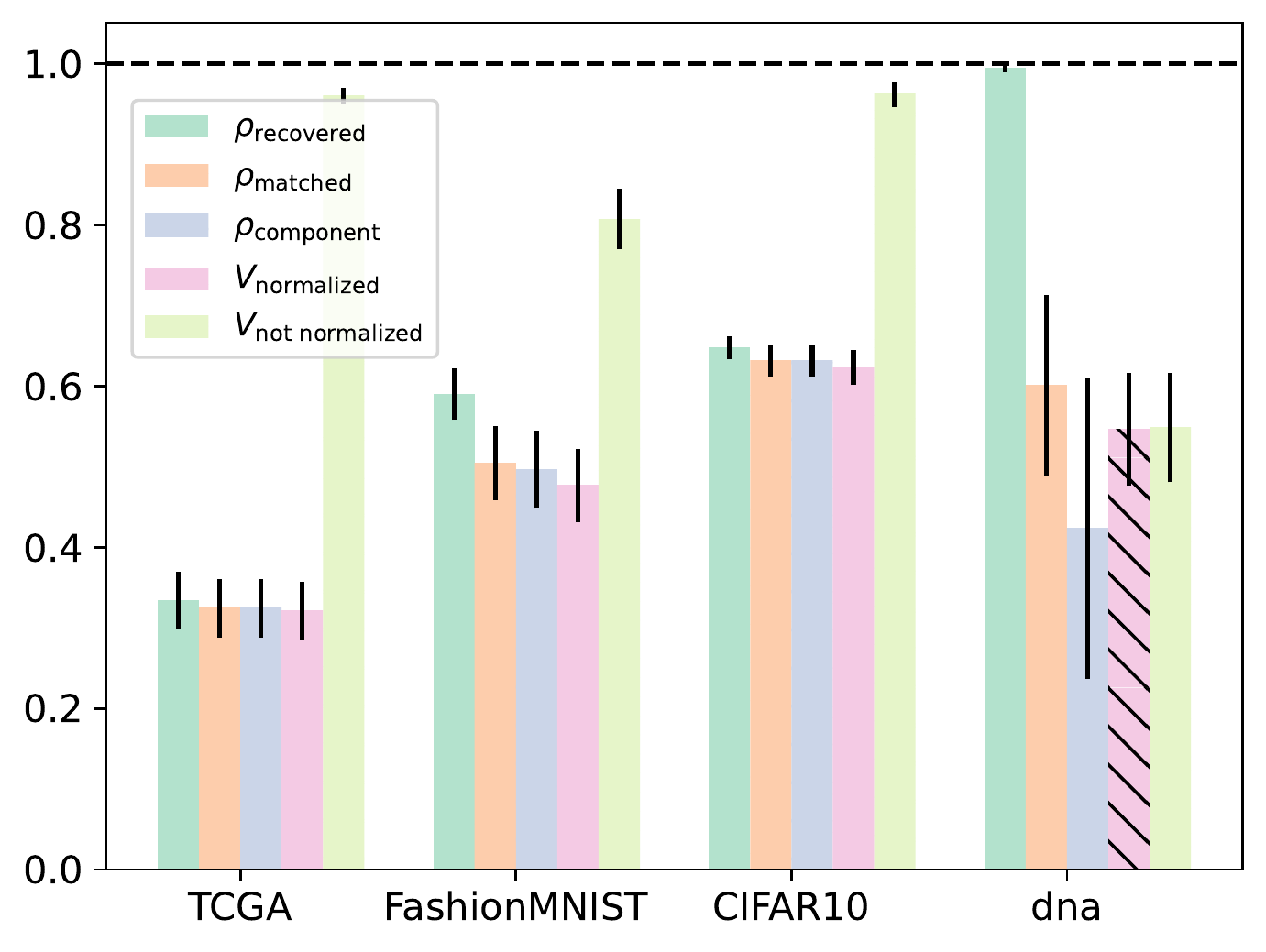}}
\caption{Result of \att\ for 4 different datasets. In each case, we consider $K=5$ clients with datasets of size $\#\mathcal{D}_k = 100$,
$n_\mathrm{updates} = 5$, $t_\mathrm{max} = 20$ with $20$ training attacks in each case. All the hyperparameters used in this figures are listed in~\Cref{tab:hp}.}
\label{fig:main_result}
\end{center}
\vskip -0.2in
\end{figure}

For each dataset, we choose realistic hyper-parameters (HP) (learning rate, number of hidden neurons), and we perform FL trainings, with $5$ clients
each containing $\#\mathcal{D}_k = 100$ data samples and with $n_\mathrm{updates} = 5$, $t_\mathrm{max} = 20$, as described in
\Cref{subsec:attack_environment} and~\Cref{alg:fedavg}.
 We then perform \att\ using only the model parameters at each round $\theta_t$ and report the metrics $\rrecov$, $\rmatched$, $\rcomponent$ and $\vweighed$ in~\Cref{fig:main_result}.

\noindent \textbf{Impact of the hyper-parameters}
In~\Cref{app:hp_results}, we explore in detail the impact of the different HP on the performance of \att\ on FashionMNIST. 
It confirms the intuition that as the number of hidden neurons, trainings or $\tmax$ grows,
so does the performance. It also demonstrates that increasing the batch size (1-32) or the number of updates (1-20) 
significantly degrades the quality of \att. However, simply increasing these parameters would not constitute 
a proper defense, as the proportion of recovered samples is still significant (> 0.1). Finally, we show that there is 
an optimal set of learning rates for which the grouping part of the attack works particularly well.
This set of learning rates does not necessarily contain the optimal learning rate in terms of performance, but is 
close enough so that it will be tested during an HP search. 
\subsection{Dependence on inter-client datasets' homogeneity}\label{sec:heterogeneity}
In this section, we explore the fact that \textit{a priori}, our grouping method is not sensitive to inter-client datasets' 
homogeneity, as it relies on a combinatorial argument. Indeed its is harder to group samples when clients' datasets are similar (homogeneous).
In order to test this hypothesis we use Dirichlet sampling~\cite{hsu2019measuring} on samples' labels to build our clients
with varying heterogeneity by varying the $\alpha$ parameter of the Dirichlet distribution (cf \Cref{app:dirichlet}). 
We compare \att\ with a 
naïve attack which consists in performing a $K$-means algorithm, $K$ being the number of clients,
on \textit{the recovered data samples $\rcal$ (from Step 1)}.

\begin{figure}[ht]
\begin{center}
\centerline{\includegraphics[width=\columnwidth]{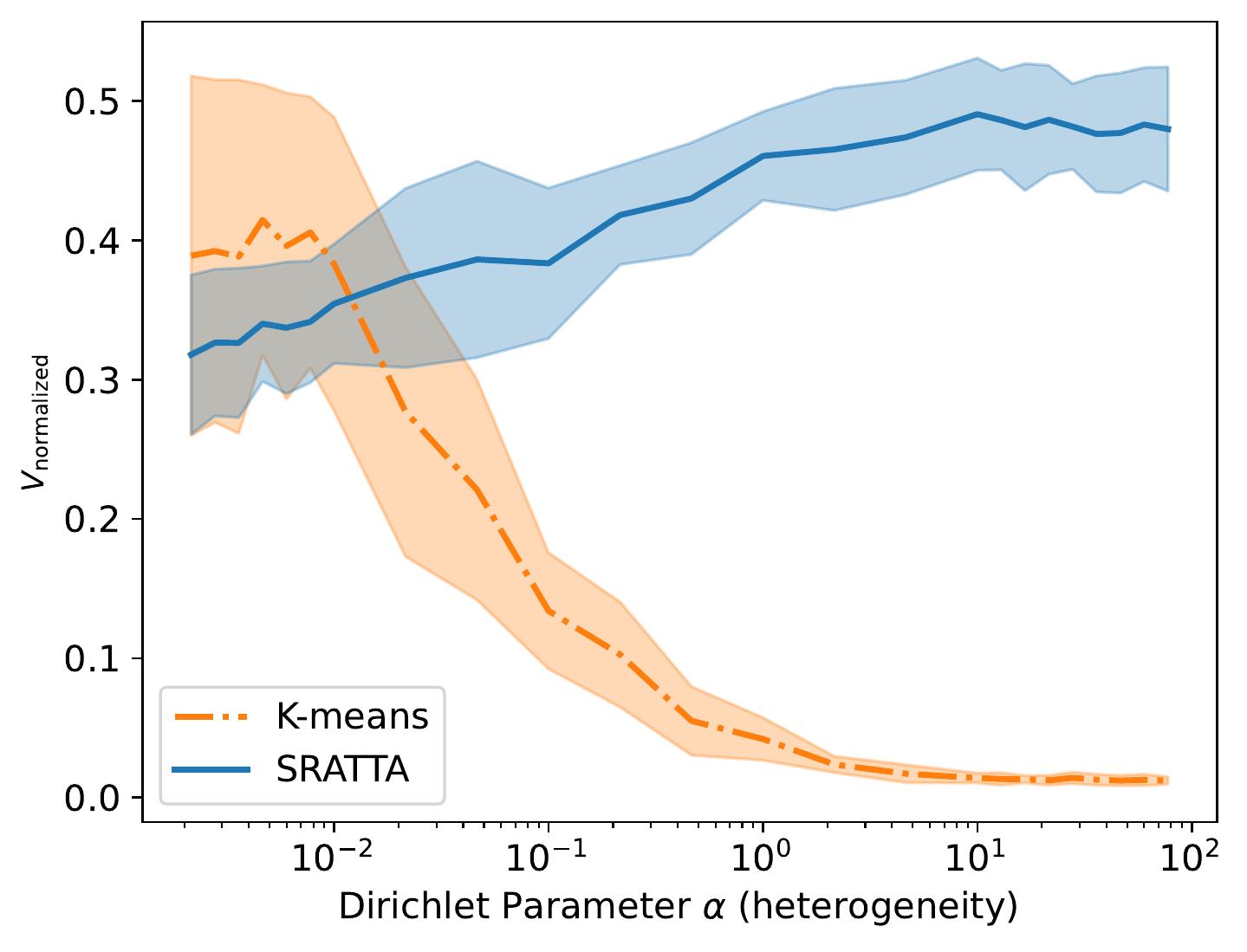}}
\caption{Success rate of \att\ and $K$-means as a function of inter-client homogeneity measured by the Dirichlet parameter $\alpha$. Cf~\Cref{app:dirichlet} and~\Cref{tab:hp} for experimental details}
\label{fig:dirichlet}
\end{center}
\vskip -0.2in
\end{figure}

We compare both methods using the normalized $V$ measure as defined above, which is particularly interesting as it takes 
into account both the homogeneity, which favours \att, as well as completeness, which favour $K$-means. 
We report the results in Fig.\ref{fig:dirichlet}. In this figure, each point of the curve corresponds to a different dataset whose labels are more homogeneously distributed amongst clients as $\alpha$ increases. $\alpha$ is not a hyper-parameter chosen by the attacker or by the clients, but a parametrization of intrinsic data heterogeneity.

The $K$-means attack works well in the heterogeneous case, and not at
all in the homogeneous case, while \att\ works reasonably well in all cases. 
\section{Fighting secure disaggregation}\label{sec:defense}
To defend against this new type of attack, we believe that clients should take matters into their own hands,
actively controlling the updates shared to the server.

We propose the following defense mechanism. At each round $t$, each client $k$ dynamically checks its own activation set $\act^h_{t,k}$ of each neuron $h$ (for a formal definition, see \Cref{eq:df_activation_client} in \Cref{app:add_def_defense}). If this activation set is too large to be attacked (larger than a threshold $q$), the corresponding update is shared. Otherwise, the neuron $h$ is frozen:
\begin{equation}
   \text{If} \quad \#\act^{h}_{t,k} \leq q,\quad \Wbar^h_{t,k,\nupdates} \text{ is reset to }  \Wbar_{t-1}^h.
\end{equation}
This removes easy-to-recover activation sets with less than $q$ samples and nullifies the effect of \att.
This censoring might reduce the final performance of the model. However in~\cref{tab:def}, we 
show that we are able to effectively defend ourselves against \att\ while preserving the accuracy of the model. In~\Cref{app:add_def_defense},~\Cref{fig:defense_acc_and_prun_all_datasets_1,fig:defense_acc_and_prun_all_datasets_2} we provide more evidence that
this defense does not affect model accuracy independently of the learning rate.
This result is coherent with related works on gradient pruning~\cite{sattler2019sparse} or dropout~\cite{srivastava2014dropout} where
models are efficiently trained with a higher proportion of frozen neurons than ours. 

We observe situations where neurons were activated by several samples but only one of them with significant
amplitude, allowing reconstruction. Moving from $q=1$ to $q=4$ improves the resilience of the defense to this case.

Finally, note that the grouping part of \att\ is relevant only if $n_\mathrm{updates} >1$. Enforcing $\nupdates =1$ effectively breaks sample matching, but it is out of the scope of this paper, for reasons detailed in~\Cref{app:def_n_update_1}.

\begin{table}[h]
   \centering\resizebox{\columnwidth}{!}{
   \begin{tabular}{lccc|c}
   \toprule
     Dataset  & $\rrecov \downarrow$  & $\vweighed \downarrow$  & $P_\mathrm{censored}$   & Model Acc. $\uparrow$  \\
   \hline 
   CIFAR10 ($q=0$)      & $0.585 \pm 0.015$       & $0.490 \pm 0.019$         & $0.000 \pm 0.000$     & $0.261 \pm 0.039$ \\
   CIFAR10 ($q=1$)      & $0.002 \pm 0.002$        & $0.002 \pm 0.002$         & $0.098 \pm 0.011$     & $0.272 \pm 0.033$ \\
   CIFAR10 ($q=4$)      & $\boldblue{0.000 \pm 0.000}$        & $\boldblue{0.000 \pm 0.000}$         & $0.180 \pm 0.014$     & $\boldsymbol{0.274 \pm 0.040}$\\
   \midrule
   DNA ($q=0$)          & $0.516 \pm 0.080$        & $0.233 \pm 0.032$         & $0.000 \pm 0.000$     & $0.929 \pm 0.017$ \\
   DNA ($q=1$)          & $0.035 \pm 0.015$        & $0.031 \pm 0.013$         & $0.005 \pm 0.002$     & $0.929 \pm 0.019$ \\
   DNA ($q=4$)          & $\boldblue{0.000 \pm 0.001}$        & $\boldblue{0.000 \pm 0.001}$        & $0.038 \pm 0.006$     & $\boldsymbol{0.932 \pm 0.020}$ \\
   \midrule
   FashionMNIST ($q=0$) & $0.476 \pm 0.027$       & $0.284 \pm 0.015$         & $0.000 \pm 0.000$       & $\boldsymbol{0.732 \pm 0.053}$ \\
   FashionMNIST ($q=1$) & $0.009 \pm 0.004$        & $0.006 \pm 0.004$         & $0.042 \pm 0.006$     & $0.729 \pm 0.043$ \\
   FashionMNIST ($q=4$) & $\boldblue{0.000 \pm 0.000}$        & $\boldblue{0.000 \pm 0.000}$         & $0.149 \pm 0.022$     & $0.731 \pm 0.046$ \\
   \midrule
   TCGA ($q=0$)         & $0.279 \pm 0.016$       & $0.193 \pm 0.010$         & $0.000 \pm 0.000$     & $\boldsymbol{0.660 \pm 0.065}$ \\
   TCGA ($q=1$)         & $0.051 \pm 0.016$        & $0.036 \pm 0.012$         & $0.135 \pm 0.016$     & $0.658 \pm 0.071$ \\
   TCGA ($q=4$)         & $\boldblue{0.000 \pm 0.000}$        & $\boldblue{0.000 \pm 0.000}$        & $0.257 \pm 0.030$     & $0.658 \pm 0.071$ \\

   \bottomrule
   \end{tabular}}
   \caption{Influence of the defense on the efficiency of \att\ and on training performance. In this table we take the point of view of the \textbf{defender/client}.  The defense strategy must therefore i) not impact the model accuracy compared to the baseline without defense and ii) nullify the attack. $P_\mathrm{censored}$ is the average proportion of neurons censored per update on the total number of neurons. We simulate an attack performed on clients defending themselves while conducting a grid search where they test 20 different learning rates. This experiment is repeated 10 times. The model accuracy reported is the best one across learning rates, averaged over repetitions. This setting highlights the defense success to prevent the attack, with almost no loss of accuracy. 
}%
   \label{tab:def}
   \end{table}
\subsection{Strengths and limitations of the defense}
As in most adversarial settings, the defense mechanism we present in this section is not proved to be absolute. Apart from the empirical evidence presented in \Cref{fig:defense_acc_and_prun_all_datasets}, two remarks strengthen our belief that knowing the defense mechanism does not result in a straightforward workaround 
for the attacker.
\begin{itemize}
\item If the attacker knows that certain neurons have been censored, retrieving them is not obvious, as this censoring is done at client level, and is not directly reflected in the aggregation.
\item Even assuming that the attacker is able to trace which neuron has been censored by which client, they can only infer that the data point which trigerred the defense lies in a certain half space (and this is assuming the knowledge of the weights defining the half space, which is not obvious considering they are modified by local updates). 
\end{itemize}
However, this does not provide a formal proof of the inviolability of the defense. 
Another point of concern could be that the defense could impact model accuracy, in the case 
where underrepresented samples are censored too often, and which is not reflected in the datasets we studied.
To mitigate this concern, we note that the defense only censors the update for specific neurons for specific clients. Thus a neuron censored by Client A can be updated by Client B in this very round, and thus not be censored at another round for the same sample. Moreover, for a sample to have virtually no impact on any neuron during the training, all neurons activated by the underrepresented sample need to be censored. 

\section{Discussion}
We believe \att\ to be an important proof of concept opening new interesting research avenues in both the design of attacks, and of privacy
layers in cross-silo FL settings.
In particular, it stresses the fact that clients have to play a bigger role in actively defending themselves, 
and that relying on secure centralized computations alone is not enough.
In our work, the restriction to models with a fully connected first layer, such as MLPs, limits the impact of \att. 

\noindent \textit{Extension to CNNs.}
Assessing if this attack can be extended to 
a broader class of models, such as CNNs, will be important 
to make sure that clients implement the appropriate defense mechanisms. 
In~\citet{pan2022exploring,zhu2021RGAP,boenisch2021curious,kariyappa2022cocktail}, the attack on MLPs is extended to CNNs by attacking the first fully connected layer. The key component needed for SRATTA to work on this setting, is a prior on the deep features allowing to filter out reconstruction candidates relying solely on their feature representations as is done now in SRATTA using raw data. Previous work \citet{pan2022exploring,zhu2021RGAP,dong2021deep} have shown that inverting the convolutional part of the network is possible, which could allow us to use priors on raw data space, and select reconstructed deep features based on their inversion to the raw data space. This is but one way to approach the problem, and we are convinced subsequent works will be able to bridge this gap.

\section*{Acknowledgements}
The authors thank the anonymous reviewers, ethics reviewer, and meta-reviewer for their feedback
and ideas, which significantly improved the paper. The authors also thank Geneviève Robin
, Alex Nowak and John Klein for their help in proofreading the paper as well as
Mathieu Andreux for his involvment in initial discussions. The authors are supported by Owkin, Inc.

\bibliographystyle{unsrtnat}
\bibliography{bibli}

\newpage
\appendix
\onecolumn
\appendix
\section{Proofs of the theorems}%
\label{app:proofs_th_activation}
\subsection{Proof of~\Cref{prp:gradient_activations}}
We provide here a proof of~\Cref{prp:gradient_activations}, which was already stated, with other notations in various works such as~\citet{aono2017privacy, geiping2020inverting, pan2022exploring}.
We start by computing the gradient of the loss at a single data point, before proving the proposition on a batch, and finally extending it to a loss which is not decomposable 
on the points of the batch (this will be useful for the Cox loss in the TCGA dataset).

In order to make differentials more explicit, we write $m(x;\theta) \defeq m_{\theta}(x)$ for $x \in \R^d,z\in\R^H,\phi\in \R^q$ and $\theta \in \R^p$.

\paragraph{1. Gradient of the loss at a single point.} Let $(x,y) \in \R^d \times \yy$.
We recall that the neural network considered starts with a fully connected layer, and hence
 \begin{equation*}
m(x;\theta)= f_{\phi}(z(x)),~z(x) \defeq \relu(Wx+b)
 \end{equation*}

For any $1 \leq h \leq H$, denote 
with $z^h$ the activation of the $h$-th hidden neuron: $z^h(x) \defeq \relu(W^h x+b^h)$. Note that $z = (z^h)_{1 \leq h \leq H}$. Using the chain rule, it holds

\eqal{eq:single_pint_activation}{
\nabla_{W^h} \ell(m(x;\theta),y) &= \frac{\partial \ell(f_{\phi}(z(x)),y)}{\partial z^h}\frac{\partial z^h(x)}{\partial W^h},~\\
\nabla_{b^h} \ell(m(x;\theta),y) &= \frac{\partial \ell(f_{\phi}(z(x)),y)}{\partial z^h}\frac{\partial z^h(x)}{\partial b^h}.
}

Note that in the main text (cf~\Cref{eq:activation_set_batch}), we define
\begin{equation} \label{eq:def_partial_loss_z}
\frac{\partial \LL(\theta;x,y)}{\partial z^h} \defeq  \frac{\partial \ell(f_{\phi}(z(x)),y)}{\partial z^h}
\end{equation}

Therefore,

\eqal{eq:single_point_activation}{
\nabla_{W^h} \ell(m(x;\theta),y) &= \frac{\partial \LL(\theta;x,y)}{\partial z^h}\frac{\partial z^h(x)}{\partial W^h},~\\
\nabla_{b^h} \ell(m(x;\theta),y) &= \frac{\partial \LL(\theta;x,y)}{\partial z^h}\frac{\partial z^h(x)}{\partial b^h}.
}

Now if $W^h x_i+b^h \leq 0$, then
$\frac{\partial z^h(x)}{\partial W^h}= 0$ and $\frac{\partial z^h(x)}{\partial b^h}=0$, taking the convention that the gradient of the $\relu$ is $0$ at $0$.

On the other hand, if $W^h x+b^h> 0$, then the gradient of the $\relu$ is $1$ and $\frac{\partial z^h(x)}{\partial W^h} = x$, $\frac{\partial z^h(x)}{\partial b^h} = 1$. In both cases, we have 

\eqal{eq:single_pooint_activation}{
\nabla_{W^h} \ell(m(x;\theta),y) &= \lambda(\theta;x,y)~x,~\\
\nabla_{b^h} \ell(m(x;\theta),y) &=  \lambda(\theta;x,y).
}
with $\lambda(\theta;x,y)  =\frac{\partial \LL(\theta;x,y)}{\partial z^h}  \mathbf{1}_{W^h x+b^h> 0}$.

\paragraph{2. Gradient of the loss on the whole batch.} Now, let $D = (x_i,y_i)_{1 \leq i \leq b} \in \R^b$ be a batch of size $b$.

The loss computed over a batch $D$ is the following sum:
\begin{equation} \label{eq:app_def_loss}
\LL(\theta;D) = \sum_{i=1}^b{\ell(m(x_i;\theta),y_i)}
\end{equation}

and hence, by applying the gradient and using~\Cref{eq:single_pooint_activation}
\eqal{eq:batch_activation}{
\nabla_{W^h} \ell(m(x;\theta),y) &= \sum_{i=1}^b{\lambda(\theta;x_i,y_i)~x_i},~\\
\nabla_{b^h} \ell(m(x;\theta),y) &=  \sum_{i=1}^b{\lambda(\theta;x_i,y_i)}.
}

By definition of the activation set $\act^h(\theta,D)$ in~\Cref{eq:activation_set_batch}, and that of $\lambda$ above we have
\[\lambda(\theta;x_i,y_i) \neq 0 \text{ i.i.f. } x_i \in \act^h(\theta,D),\]

which concludes the proof. 

\paragraph{3. Gradient of a loss which is not separable on a batch } Assume now that the loss is not decomposable along points of the batch, but that we have instead
\begin{equation} \label{eq:app_def_loss_2}
\LL(\theta;D) \defeq \ell(m_{\theta}(x_D),D), \text{ where }m_{\theta}(x_D)\defeq (m_{\theta}(x_i))_{1 \leq i \leq b}\in \R^{C \times b},
\end{equation}
for some differentiable loss $\ell(\cdot,D): \R^{C \times b} \rightarrow \R$ on the batch. Then in the same way, we define the activation set as

\eqal{eq:activation_set_batch_bis}{
\act^h(\theta,D) \defeq \{x_i \in D~:~ &W^h \cdot x_i + b^h > 0 \\
			\text{ and } &\frac{\partial \LL(\theta;D)}{\partial z^h_i} \neq 0\},
}

where $\frac{\partial \LL(\theta;D)}{\partial z^h_i}$ is the differential of the loss w.r.t. the output of the first neuron at parameter $\theta$ on sample $x$, 
which is formally the partial derivative of $Z = (z_1,...,z_b) \in \R^{C \times b} \mapsto \ell((f_{\phi}(z_i))_{1 \leq i \leq b},D)$ with respect to $[Z]_{hi} = z_i^h$ at point $Z = (\relu(Wx_i + b))_{1 \leq i \leq b}$.

We can then compute the partial derivatives, and obtain exactly the same result as the one above.

\subsection{Proof of~\Cref{prp:round_activations}}
\label{proof_t}
As the clients are performing standard SGD,
the updates of the clients are a sum of the gradients computed at each update. Besides, as the FL protocol is FedAvg, the aggregated model $\theta_t$
is a simple average of all the client updates. Therefore:

\[\Delta \theta_t = -\tfrac{\eta}{K} \sum_{k=1}^K{\sum_{i=0}^{\nupdates-1}{\nabla \LL(\theta_{t,k,i},B_{t,k,i})}}\]
And thus for the first layer's weights and biases we have:
\[\Delta W^{h}_t = -\tfrac{\eta}{K} \sum_{k=1}^K{\sum_{i=0}^{\nupdates-1}{\sum_{j=0}^{b} \lambda(\theta_{t,k,i}; x_j^{B^{i}_{k}}, y_j^{B^{i}_{k}}) x_j}}\]
\[\Delta b^{h}_t = -\tfrac{\eta}{K} \sum_{k=1}^K{\sum_{i=0}^{\nupdates-1}{\sum_{j=0}^{b} \lambda(\theta_{t,k,i}; x_j^{B^{i}_{k}}, y_j^{B^{i}_{k}})}}\]
Where $x_j^{B^{i}_{k}}$ indicates the $j$-th sample of the batch of client $k$ at update $i$.

Therefore,~\Cref{prp:round_activations} is a straightforward consequence of~\Cref{prp:gradient_activations}.

\subsection{Proofs of~\Cref{th:mainth}}
\label{app:proof_main_th}
Let $x \in \act^h_t$. Assume that $x$ is a sample from client $k$; since we suppose that~\Cref{asm:not_common_data} is satisfied, then $x$ must be in one of the batches 
of client $k$, \textit{i.e.}, must belong to one of the batches in the sequence $(B_{t,k,i})_{0 \leq i \leq \nupdates -1}$. Note that if~\Cref{asm:not_common_data} were not satisfied, it could 
belong to another client's batches. 

Let $i^h_k$ be the minimal index of a batch in the sequence $(B_{t,k,i})_{0 \leq i \leq \nupdates -1}$ containing a sample which activates neuron $h$, (it is well defined because $x$ is such an element),
 that is 
\[i_k = \min \{0 \leq i \leq \nupdates ~:~\act^h(\theta_{t,k,i},B_{t,k,i}) \neq \emptyset \}.\]

Since for all $i< i^h_k$, we have $\act^h(\theta_{t,k,i},B_{t,k,i}) = \emptyset$, by~\Cref{prp:gradient_activations}, we have $\nabla_{\Wbar^h}{\LL(\theta_{t,k,i},B_{t,k,i}) } = 0,~0 \leq i < i^h_k$.
Using the fact that SGD updates are performed, $\Wbar^h_{t,k,i+1} = \Wbar^h_{t,k,i} - \eta \nabla_{\Wbar^h}{\LL(\theta,B_{t,k,i}) } = \Wbar^h_{t,k,i}$ for all $0 \leq i < i^h_k$, meaning that 
 \[\Wbar^h_{t-1} = \Wbar^h_{t,k,0} = ... =  \Wbar^h_{t,k,i^h_k}.\]
 
Let $x'$ be an element in $\act^h(\theta_{t,k,i},B_{t,k,i^h_k})$ (it is possible that $x'$ is in fact $x$). 
It is in $\act^h_t$ by definition, and since $\Wbar^h_{t,k,i_k^h} = \Wbar^h_{t-1}$, it is also in $\actbis$. Thus, we have proven
that there is an element $x'$ from client $k$ in $\actbis$.

\subsection{Proof of~\Cref{cor:maincor}}
\label{app:proof_corr}
Assume all elements of $\actbis$ come from the same client $k$.
Let's consider $x \in \act^h_t$. According to~\Cref{th:mainth}, there exists $x' \in \actbis$ such that $x$ and $x'$ are from the same client.
As $x'$ comes from client $k$, $x$ comes from client $k$. Therefore, all elements $x \in \act^h_t$ come
from client $k$. In particular, if $\actbis$ contains only one element, by~\Cref{asm:not_common_data}, this element belongs to one and only one client $k$. The result then follows.

\section{Recovering the activation sets using Orthogonal matching pursuit}
\label{app:OMP}

The goal of this section is to explain how we do the second step of \att\ described in \textbf{Step 2} of~\Cref{secsec:main_attack}.

\subsection{Problem reformulation}
Recall from~\Cref{eq:activation_reconstruction} that 
 for each couple $(t,h)$ of neurons and rounds, we try to find
$N \in \N$,~$(\lambda_r)_{1 \leq r \leq N}$ and $(x_r)_{1 \leq r \leq N} \in \rcal^N$ such that

\begin{subequations}%
\label{eq:activation_reconstruction_aux}
\begin{align}
\Delta W^h_t &= \sum_{r=1}^{N} \lambda_r x_r \\
\Delta b^h_t &= \sum_{r=1}^{N} \lambda_r \\
\forall r,\ &\lambda_r \neq 0
\end{align}
\end{subequations}

Let us reformulate~\Cref{eq:activation_reconstruction_aux}.
Denote with $\Ncal$ the number of recovered samples in $\rcal$ and let $x_1,..,x_{\Ncal} \in \rcal$ be the list of recovered samples. Let $X \defeq (x_1,...,x_{\Ncal})^{\top} \in \R^{N_{\rcal}\times d}$ be the matrix of recovered samples, and $\Xbar \defeq \begin{pmatrix} X & \mathbf{1} \end{pmatrix} \in \R^{\Ncal \times (d+1)}$ be the extended recovered samples matrix.~\Cref{eq:activation_reconstruction_aux} can simply be written 

\begin{equation}
\label{eq:first_problem_omp}
\text{ Find  } \lambda \in \R^{\Ncal} \text{ s.t. } \Xbar^{\top} \lambda = \Wbar^h_t,
\end{equation}

and a recovered activation set can be defined from $\lambda$ as the set of recovered samples whose corresponding coefficient $\lambda_r$ is non zero, that is $\actapprox^h_t(\lambda) \defeq \{x_r \in \rcal~:~\lambda_r \neq 0\}$. 

\subsection{Restriction to small activation sets}

Solving~\Cref{eq:first_problem_omp} and using a resulting $\lambda$ to build the recovered activation set $\actapprox^h_t(\lambda)$ is not satisfactory. Indeed, as soon as $\Ncal > d+1$ there are infinitely many solutions to~\Cref{eq:first_problem_omp}, corresponding to potentially infinitely many recovered activation set $\actapprox^h_t(\lambda)$. In that case, there is no way of knowing which one of these approximations is actually the true activation set.

As explained in the main text (see~\Cref{secsec:main_attack}), to avoid this problem, we further constrain~\Cref{eq:first_problem_omp} to the $\lambda$ corresponding to activation sets of size at most $\Nmax$, where $\Nmax$ is a hyper-parameter selected by the attacker in order to recover meaningful activation sets, and where $\Nmax \ll d$ in order to avoid the problem of having infinitely many solutions. This can be formalized as the following problem

\begin{equation}
\label{eq:second_problem_omp}
\text{ Find  } \lambda \in \R^{N_{\rcal}} \text{ such that }  \|\lambda\|_0 \leq \Nmax \text{ and } \Xbar^{\top} \lambda = \Wbar^h_t, 
\end{equation}

where $\|\lambda\|_0$ denotes the number of non zero coefficients of $\lambda$. The intuition behind this restriction is that (i) we wish to recover small activation sets and (ii) it is much less likely that~\Cref{eq:second_problem_omp} has a solution with a small $\Nmax$, so that if we find one, the corresponding $\actapprox^h_t(\lambda)$ is likely to be equal to the true activation set $\act^h_t$. 
In practice, to solve~\Cref{eq:second_problem_omp}, we solve the relaxed version

\begin{equation}
\label{eq:last_problem_omp}
\text{ Find  } \lambda \in \R^{N_{\rcal}} \text{ such that } \lambda \in \arg\min_{\|\lambda\|_0 \leq \Nmax} \left\| \Xbar^{\top} \lambda - \Wbar^h_t\right\|^2 .
\end{equation}

We then check that the recovered solution $\lambda$ satisfies $\Xbar^{\top} \lambda = \Wbar^h_t $ with sufficient precision (if it is not the case, this means that~\Cref{eq:second_problem_omp} has no solution). In the main paper, we denote with $\Lrec$ the set of couples $(t,h)$ for which~\Cref{eq:second_problem_omp} has a solution.

This last formulation is quite standard, and can be solved using different methods, such as orthogonal matching pursuit (OMP),~\cite{rubinstein2008efficient,mallat1993matching} or the LASSO method~\cite{kim2007}. In practice, we solve this problem using OMP, and $\Nmax$ is selected in order to lead to clusters of reasonable size. OMP is based on a greedy algorithm and builds the activation set iteratively. At each step, it adds to the activation set the sample which is most highly correlated with the residual. It is called orthogonal matching pursuit because at each iteration, the residual is recomputed using an orthogonal projection on the current state of the activation set. For more details, see~\citet{mallat1993matching} as well as the scikit-learn documentation~\citet{scikit-learn}.

\section{Relaxation of hypotheses}
\label{app:relaxation}
We note that although we assume full-participation in the rest of the article it is not strictly necessary for \att\
to work. In addition, the local and global learning rate, batch size, number of updates do not have to be fixed. The only important
thing is to perform multiple gradient updates, and that the learning rate is the same for the weights and the bias at
each local minibatch's gradient step. The sum across clients could be weighted (in fact this can be included in the label
variable, by adding an extra label variable $k$ for the client).
\section{Pseudocode of \att}
The pseudocode of \att\ can be found below in~\Cref{alg:pseudocode}.

   \begin{algorithm}[H]
   \caption{\att}
         \label{alg:pseudocode}
\begin{algorithmic}
   \REQUIRE number of trainings $num_{trainings}$, number of rounds per training $t_{max}$, 
   number of hidden neurons in the first layer of the model $H$, model updates for all trainings, 
   rounds, neurons $(\Delta W^h_t, \Delta b^h_t)$, a data prior on the data $\pp$, maximum number of 
   samples allowed in the Orthogonal Matching Pursuit $\Nmax$

   \begin{center}
      ---------------- Step 1: Sample recovery ----------------
   \end{center}

   \STATE $ \rcal \leftarrow \emptyset $
   \FOR{$training=1$ {\bfseries to} $num_{trainings}$}
      \FOR{$t=1$ {\bfseries to} $t_{max}$}
         \FOR{$h=1$ {\bfseries to} $H$}
            \STATE $ candidate = \Delta W^h_t / \Delta b^h_t $
            \IF{$distance(candidate, \pp) \approx 0$}
               \STATE $\rcal \leftarrow \rcal \cup \{candidate\}$
            \ENDIF
         \ENDFOR
      \ENDFOR
   \ENDFOR

   \begin{center}
      ---------------- Step 2: Reconstruction of activation sets ----------------
   \end{center}

   \STATE $ L_{rec} \leftarrow \emptyset $
   \FOR{$training=1$ {\bfseries to} $num_{trainings}$}
      \FOR{$t=1$ {\bfseries to} $t_{max}$}
         \FOR{$h=1$ {\bfseries to} $H$}
            \STATE $ \{ (\lambda_r, x_r): 1 \leq r \leq N \} \leftarrow \text{ Orthogonal Matching Pursuit}(\rcal, \Delta W^h_t, N_{max}) $
            \STATE $ \act^h_t = \{ x_r: 1 \leq r \leq N \} $
            \IF{ $ \act^h_t \neq \emptyset \text{ and } \Delta W^h_t \approx \sum_{r=1}^{N} \lambda_r x_r \text{ and } \Delta b^h_t \approx \sum_{r=1}^{N} \lambda_r $ }
               \STATE $ L_{rec} \leftarrow L_{rec} \cup (h, t) $
            \ENDIF
         \ENDFOR
      \ENDFOR
   \ENDFOR

   \begin{center}
      ---------------- Step 3a): Definitions ----------------
   \end{center}

   \FOR{$training=1$ {\bfseries to} $num_{trainings}$}
      \FOR{$ (h, t) \in L_{rec} $}
         \STATE $\actbis \leftarrow \{x \in \act^h_t ~:~  W_{t-1}^h x + b^h_{t-1} > 0\}$
      \ENDFOR
   \ENDFOR

   \STATE $ relationshiplist \leftarrow [] $
   \FOR{$training=1$ {\bfseries to} $num_{trainings}$}
      \FOR{$ (h, t) \in L_{rec} $}
            \STATE $relationshiplist.append( [ \actbis, \act^h_t \setminus \actbis ] ) $
      \ENDFOR
   \ENDFOR
   \begin{center}
      ---------------- Step 3b): Creating groups ----------------
   \end{center}
\STATE $ \mathcal{G} \leftarrow \text{ edgeless graph where each node } \in \mathcal{R} $
   \WHILE{the number of edges in $\mathcal{G}$ is growing}
      \FOR{$ relationship \in relationshiplist $}
         \IF {all samples from $ relationship[0]$ are connected}
            \STATE Draw an edge in $\mathcal{G}$ between all samples in $ relationship[0] $ and all samples in $ relationship[1] $
         \ENDIF
      \ENDFOR
   \ENDWHILE

   \STATE $clusteredclients \leftarrow \text{ connected components of } \mathcal{G} $

   \ENSURE $clusteredclients$
\end{algorithmic}
\end{algorithm}

\newpage
\section{Details on numerical experiment}

\subsection{Details on numerical experiment}\label{app:num_exp}

\paragraph{Additional details on the metrics used.}

We can see the output of \att\ as a partition of the set of recovered samples $P_1\cup ... \cup P_p = \rcal$, where the 
sets $P_1,...,P_p$ which are disjoint. The sets $P_i$ are inferred by the attacker in step 3b), and represent the largest 
possible groups of samples which they can build. We will assess the quality of the sample recovery process using the following 
different metrics.

\noindent $\rrecov \defeq \# \rcal / \#\dd$ is the ratio of recovered samples and measures the quality of the recovery process.

\noindent $\rmatched$ is the ratio of samples in the dataset which have been recovered and grouped with at least another 
sample (that is samples that belong to $P_i$ with $\# P_i > 1$).

\noindent $\rcomponent$ is the ratio between the average size of the $K$ biggest elements in the partition $(P_1,...,P_p)$ 
and the average size of a client dataset $\#\mathcal{D}_k$ (In practice, all client datasets will be of equal size). 
The intuition is that the $K$ biggest elements of the partition would ideally each correspond to a different client dataset.

\noindent $\vweighed \defeq \rrecov * \vrec$ is the normalized $V$-measure of the grouped recovered samples, where $\vrec$ is 
the $V$-measure \cite{rosenberg2007v} of the partition $(P_1,...,P_p)$ of $\rcal$ with respect to the true partition of 
$\rcal$ into $K$ sets across the $K$-clients. Exactly in the same way that the $F1$ score mixes precision and recall, here,
the $V$-measure mixes two measures of the quality of the partition $P_1,...,P_p$: homogeneity $h \in [0,1]$ and completeness
$c \in [0,1]$. Homogeneity measures the diversity of each class: $h =1$ if each $P_i$ contains elements coming from only
one client (this will be the case if we have no false positive). Completeness measures how each true client is distributed
amongst different groups $P_i$ ; $c = 1$ if for all $k$, all elements belonging to client $k$ are assigned to the same
element of the partition $P_{i_k}$. $\vrec$ is the harmonic mean of these two quantities, so that $\vrec =0$ if either
$c,h=0$ and $\vrec = 1$ if both $c,h=1$. This measure is particularly adapted to our setting as we do not have a natural
metric to compare different clusterings with a different number of clusters. We multiply by $\rrecov$ so that the attacker
clusters all data points and to the right client if and only if $\vweighed = 1$. $\rrecov$ can be seen as the recall but taken
over all samples not only recovered samples.

\subsection{Datasets used}

\label{app:datasets}
\subsubsection{CIFAR10}
CIFAR10~\cite{krizhevsky2009learning} is one of the most well-known image classification datasets. There are $60,000$ object-centric images of relatively low
resolution 32x32x3 that are labeled with 10 different mutually exclusive classes (horse, frog, truck, airplane, automobile, bird, cat, deer, dog, ship).

\subsubsection{FashionMNIST}
FashionMNIST~\cite{xiao2017/online} was released as a more challenging alternative to MNIST~\cite{lecun1998gradient} with grayscale images 28x28 of fashion items
split across 10 classes (T-shirt/top, Trouser, Pullover, Dress, Coat, Sandal, Shirt, Sneaker, Bag, Ankle boot). There are $70,000$ images in total.

\subsubsection{DNA}
Primate Splice-Junction Gene Sequences (or DNA dataset) dataset is a part of the OpenML suite~\cite{vanschoren2014openml}. It consists of $3,186$ samples which are splice junctions. 
Each sample is a vector of $180$ indicator binary variables and the ML task is a 3-way classification task (ei, ie, neither). More details
can be found in~\cite{vanschoren2014openml}.

For all the datasets above, we use MLP networks directly on raw pixels that are just rescaled to be between 0 and 1 (with 255 distinct values).

\subsubsection{TCGA-BRCA}

The Cancer Genome Atlas (TCGA)~\cite{weinstein2013cancer, Tomczak2015} is an open initiative to gather and make available oncology patients' data
for multiple modalities such as tabular, radiology, histology and genomics and different cancers. The TCGA is organized
by study where each study focuses on a single cancer. BRCA is the BReast CAncer study. The samples in TCGA were collected
across multiple institutions that are tagged by barcodes and identifiable, this is also the case for BRCA. 
FLamby~\cite{terrail2022flamby} uses such naturally split data inside TCGA-BRCA to provide a realistic distributed dataset
where the ML task is survival prediction~\cite{jenkins2005survival}. For a more complete treatment of the
ML task defined by this dataset see Appendix F in~\cite{terrail2022flamby}. We use this tabular dataset
with $1088$ patients and retrieve for each patient a feature obtained from its histology slides (see \textit{Primer on histology}
below if needed). We detail below in \textit{Histology Preprocessing} the exact process used to extract such features and join both datasets.  

\textbf{TCGA license}  

The data terms can be found \href{https://gdc.cancer.gov/access-data/data-access-processes-and-tools}{https://gdc.cancer.gov/access-data/data-access-processes-and-tools}. 
In particular, as per the GDC data access policy, users should not attempt to identify individual human research participants
from whom the data were obtained.

\textbf{Primer on histology}  

Histology slides~\cite{ghaznavi2013digital} are large-size images obtained by digitizing stained slices of biopsies of
the patient's tumor sites. Such images, called Whole Slide Images (WSI), generally comprises millions of pixels
representing the slice at different resolution and comprise a large uninformative background. This background has to be
removed to maximize the Peak-Signal-to-Noise-Ratio of the information inside the WSI. As this data usually does not fit in 
GPU RAM, its dimensions are reduced using tiling and feature extraction~\cite{courtiol2018classification}.  

\textbf{Histology Preprocessing}  

Like in~\cite{andreux2020federated}, we use natural splits and retrieve the original histology sides corresponding
to each patient using the patient ID. As some patients have multiple histology slides we keep only the first one after alphabetical
sorting. We drop all patients for which histology slides could not be found ($38$ in total making a final total of $1050$ patients). 
We then tile the matter on each slide at x20 magnification using a U-net~\cite{ronneberger2015u} trained on an in-house
dataset and compute ResNet-50 features~\cite{he2016deep} pretrained on IMAGENET~\cite{deng2009imagenet} for each extracted
tile. Therefore each patient is associated with a row of clinical data and a variable number of ResNet-50 features proportional
to the amount of matter found on the corresponding slide. We then train a linear autoencoder with bottleneck size $256$
on the ResNet-50 features from all slides using an L2 cost as is common practice~\cite{schmauch2020deep}. Once this autoencoder
is trained, we encode all features in each slide. Finally, we average all autoencoded features on each slide to get a
fixed-sized feature, following MeanPool architectures~\cite{saillard2021self}. Therefore for each patient we build a feature
vector of size $39+256=295$ with $39$ clinical features and $256$ "histology"
features.

\textbf{Clinical Data Preprocessing}
We use FLamby\cite{terrail2022flamby}'s preprocessed version of TCGA-BRCA where original features are filtered and binarized
except for the age feature (of integer type), which is left unaltered. This allows us to easily define the corresponding data prior
to filter out candidates. See~\cite{terrail2022flamby}'s Appendix F1 and~\cite{andreux2020federated} for more details about
the exact preprocessing steps.

\textbf{Survival Loss function}

One of the foundation models in survival analysis is the linear Cox proportional hazard~\cite{cox1972regression}. This model assumes:
\begin{equation}
h(t,x) = h_0(t) \exp(\beta^T x)
\end{equation}
where $h_0$ is the baseline hazard function (common to all patients and dependent on time only) and $\beta$ is the vector of parameters of our linear model.
$\beta$ is estimated by minimization of the negative Cox partial log-likelihood, which compares relative risk ratios:
\begin{equation}
L(\beta) = - \sum_{i: \delta_i = 1} \Big[ \beta^T x_i - \log \big( \sum_{j: t_j > t_i} \exp( \beta^T x_j ) \big) \Big]
\end{equation}
where $i$ and $j$ index patients and $\delta_i = 1$ indicates an event. This loss is extended in our case to handling $m_{\theta}(x_i)=\relu(Wx_i + b)$ instead of raw $x_i$
where we backpropagate through $W$ and $b$ as well. This is a natural extension explored in~\cite{andreux2020federated}:
\begin{equation}
   L(\beta) = - \sum_{i: \delta_i = 1} \Big[ \beta^T m_{\theta}(x_i) - \log \big( \sum_{j: t_j > t_i} \exp( \beta^T m_{\theta}(x_j) ) \big) \Big]
\end{equation}
We minimize the negative Cox partial log-likelihood by mini-batch gradient descent w.r.t. $\beta$.
We note that, as this loss is non-separable, its gradient on mini-batches might be degenerate if the sampled batch contains only non-admissible
pairs that cannot be ranked together~\cite{jenkins2005survival}.

All experiments in the article on the presented datasets are repeated $10$ times in order to produce confidence intervals.

\subsection{Dirichlet split of the dataset across the clients.}%
\label{app:dirichlet}
In~\Cref{sec:heterogeneity} and \Cref{fig:dirichlet}, we use the same method as~\cite{hsu2019measuring} to split the FashionMNIST dataset into different clients,
while parameterizing the heterogeneity of the data distribution by a parameter $0 < \alpha < \infty$. A small value of $\alpha$ corresponds to a heterogeneous data distribution, while a high value of $\alpha$ corresponds to a homogeneous data distribution. 

More precisely, for a given value of $\alpha$ and a given client $k$, we draw a random vector $\vec q$ following a Dirichlet distribution of parameter $\alpha$ of size $L=10$, where $L$ is the number of labels present in FashionMNIST. Such a vector satisfies $\|\vec q\|_1 = 1$ and $q_l \geq 0$ for $l = 1, \dots, L$. The density distribution of $\vec q$ is proportional to $\prod_l q_l^\alpha$. For the client $k$, we then generate a dataset of size $\#\dd_k$, such that the number of samples with label $l$ is roughly $\lfloor q_l \#\dd_k\rfloor$.

When $\alpha \to \infty$, we have $q_l \to 1/L$ for all $l$, therefore each client $k$ as $\#\dd_k / L$ samples of each label. When $\alpha \to 0$, we have $\vec q \to (0., \dots, 0, 1, 0, \dots)$, where the only non-zero component of $\vec q$ is randomly one of the $L$ dimensions. In that case, each client has samples coming from only one label.

\begin{figure}[ht]
\begin{center}
\centerline{\includegraphics[width=\columnwidth]{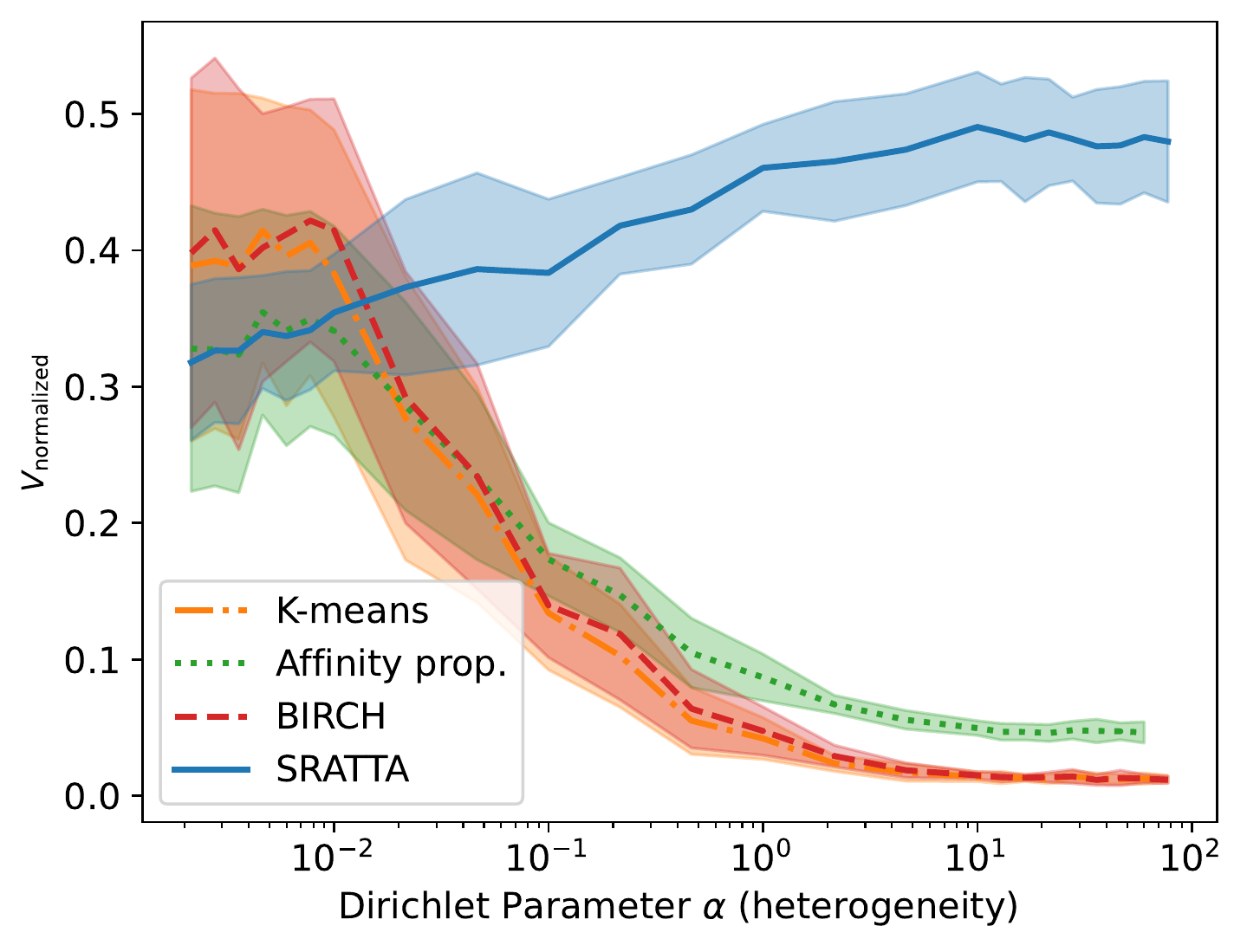}}
\caption{Success rate of \att\ and $K$-means, affinity propagation \cite{frey2007clustering} and BIRCH \cite{zhang1996birch} as a function of inter-client homogeneity measured by the Dirichlet parameter $\alpha$. Cf~\Cref{app:dirichlet} and~\Cref{tab:hp} for experimental details}
\label{fig:dirichlet_extended}
\end{center}
\vskip -0.2in
\end{figure}

We provide in \Cref{fig:dirichlet_extended} an extension of \Cref{fig:dirichlet} where we added the performance of two extra clustering algorithms: the BIRCH \cite{zhang1996birch} and the Affinity propagation \cite{frey2007clustering}. We could not empirically make the attack works with other standard clustering algorithms.

\subsection{Effect of different hyperparameters on the strength of \att}
\label{app:further_result}

\begin{figure}[h]
   \centering
   \begin{subfigure}[b]{0.4\textwidth}
      \centering
      \includegraphics[width=\textwidth]{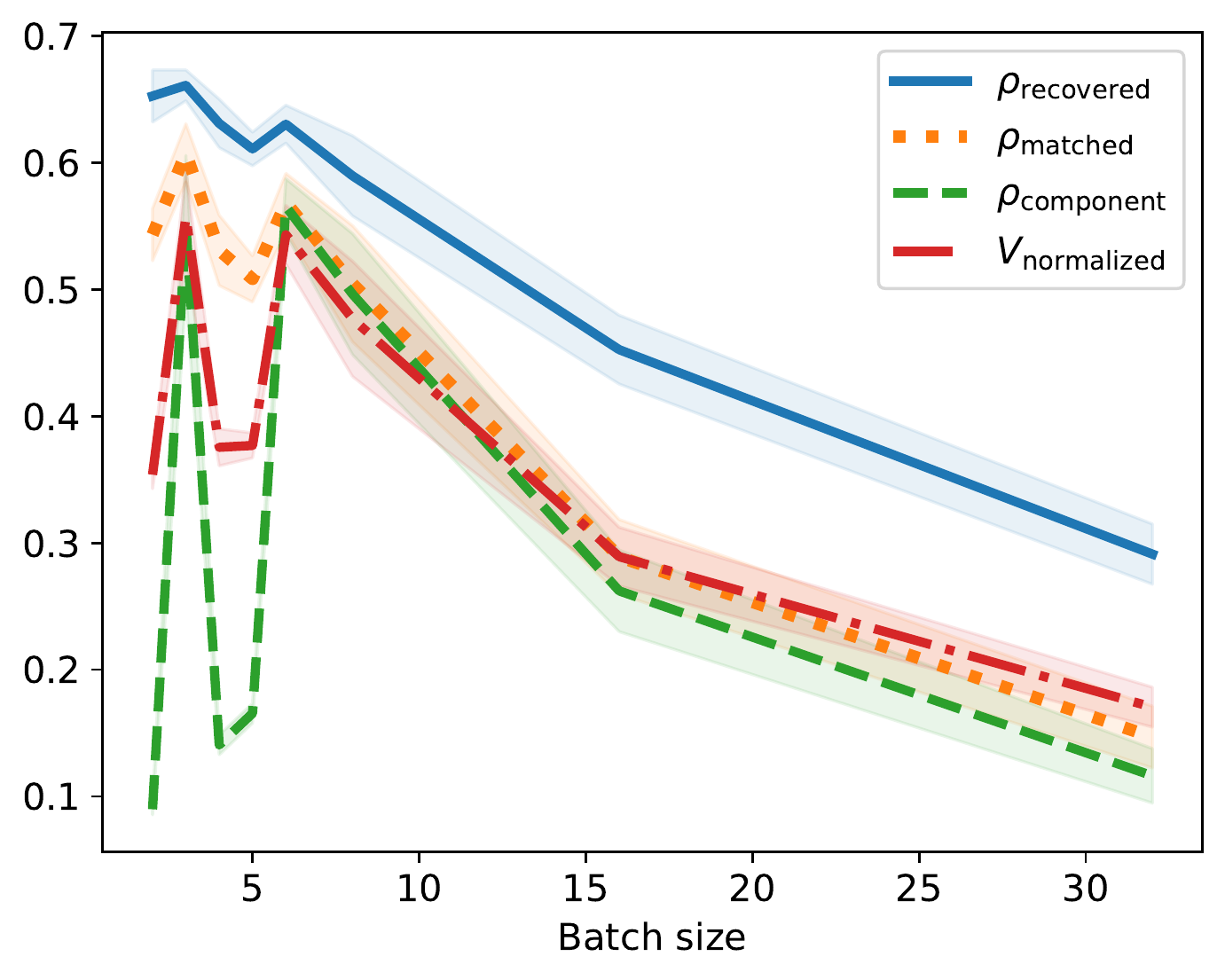}
      \caption{}
      \label{fig:effect_bs}
   \end{subfigure}
   \begin{subfigure}[b]{0.4\textwidth}
      \includegraphics[width=\textwidth]{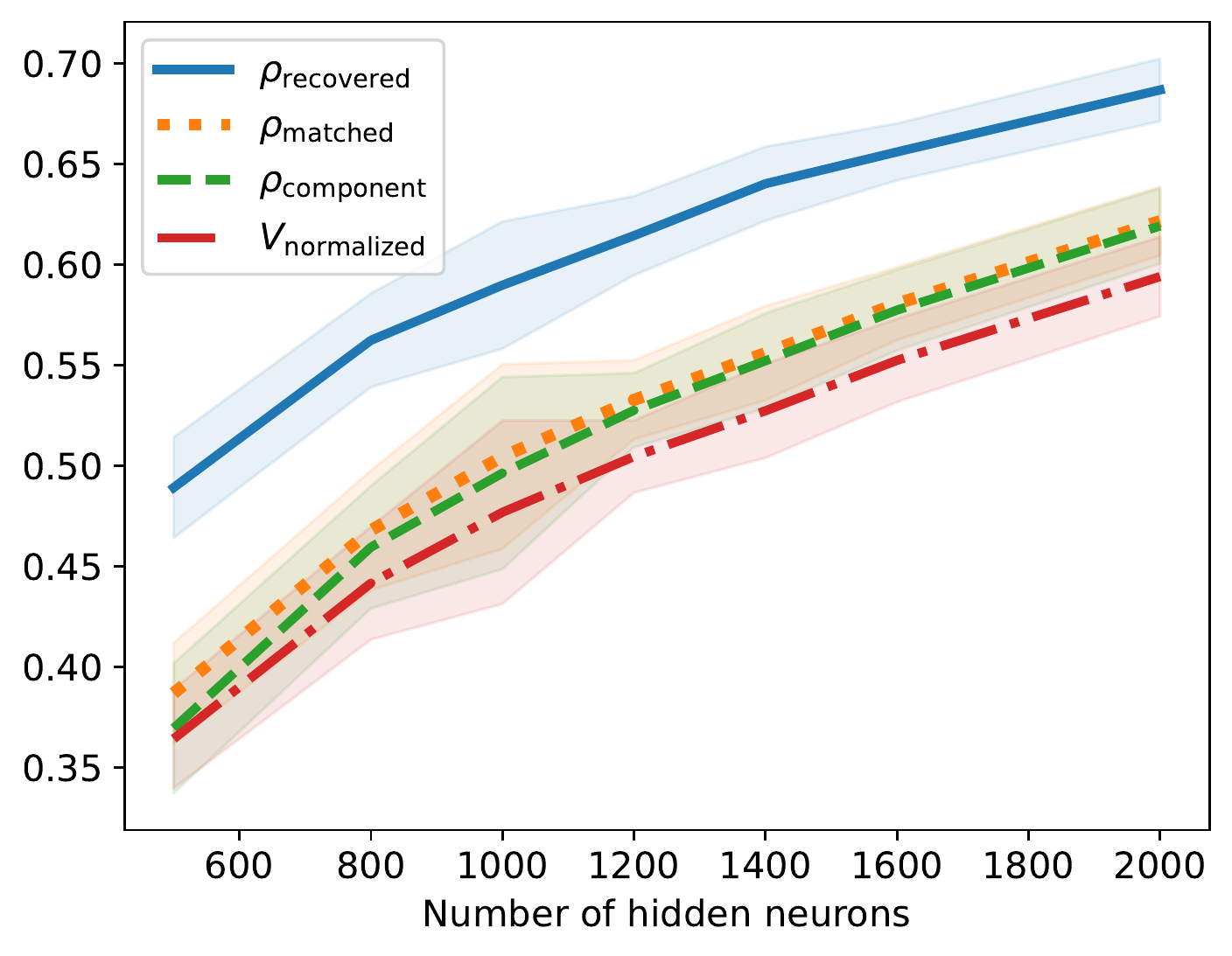}
            \caption{}
      \label{fig:effect_neurons}
   \end{subfigure}
      \begin{subfigure}[b]{0.4\textwidth}
      \includegraphics[width=\textwidth]{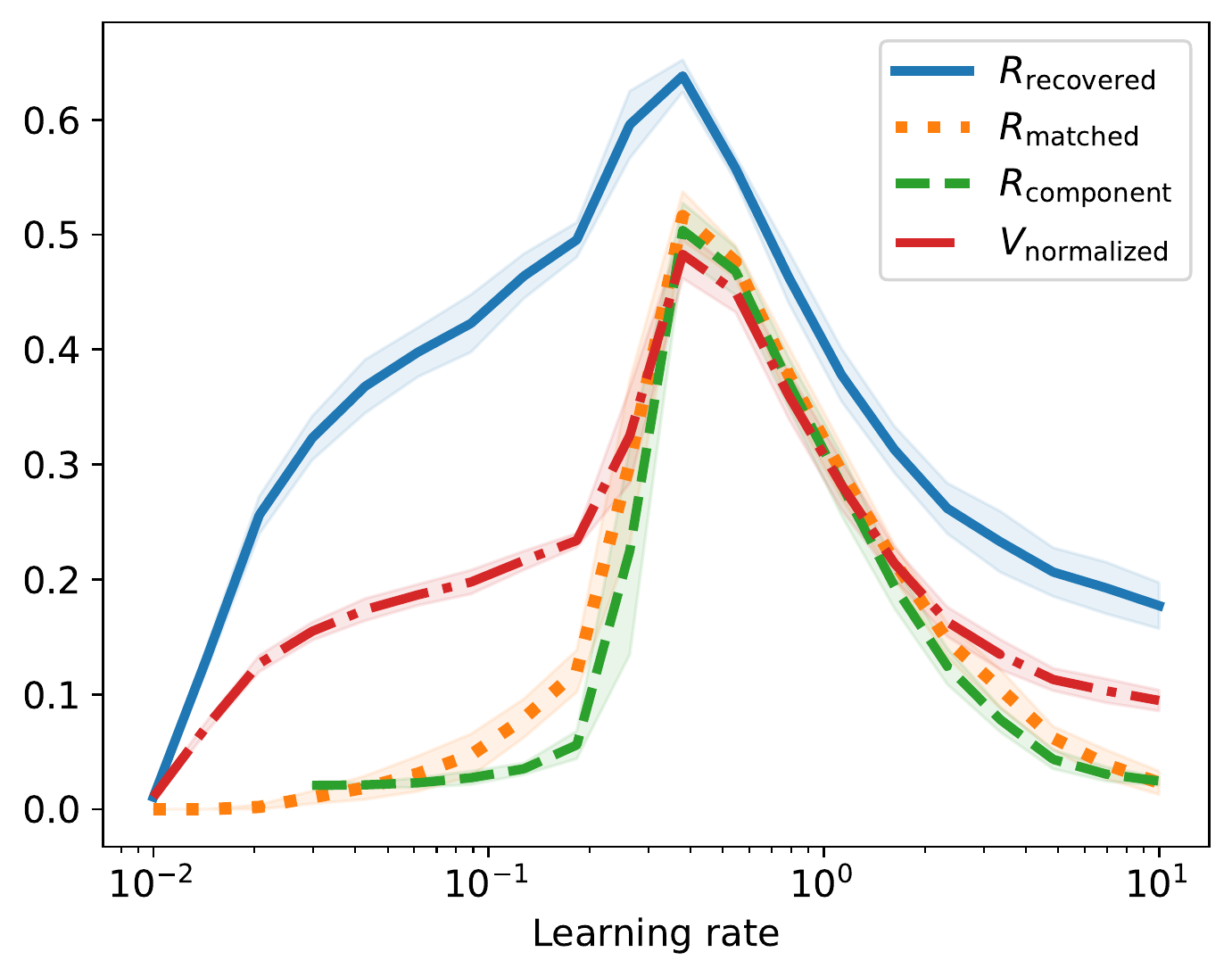}
            \caption{}
      \label{fig:effect_lr}

   \end{subfigure}
      \begin{subfigure}[b]{0.4\textwidth}
      \includegraphics[width=\textwidth]{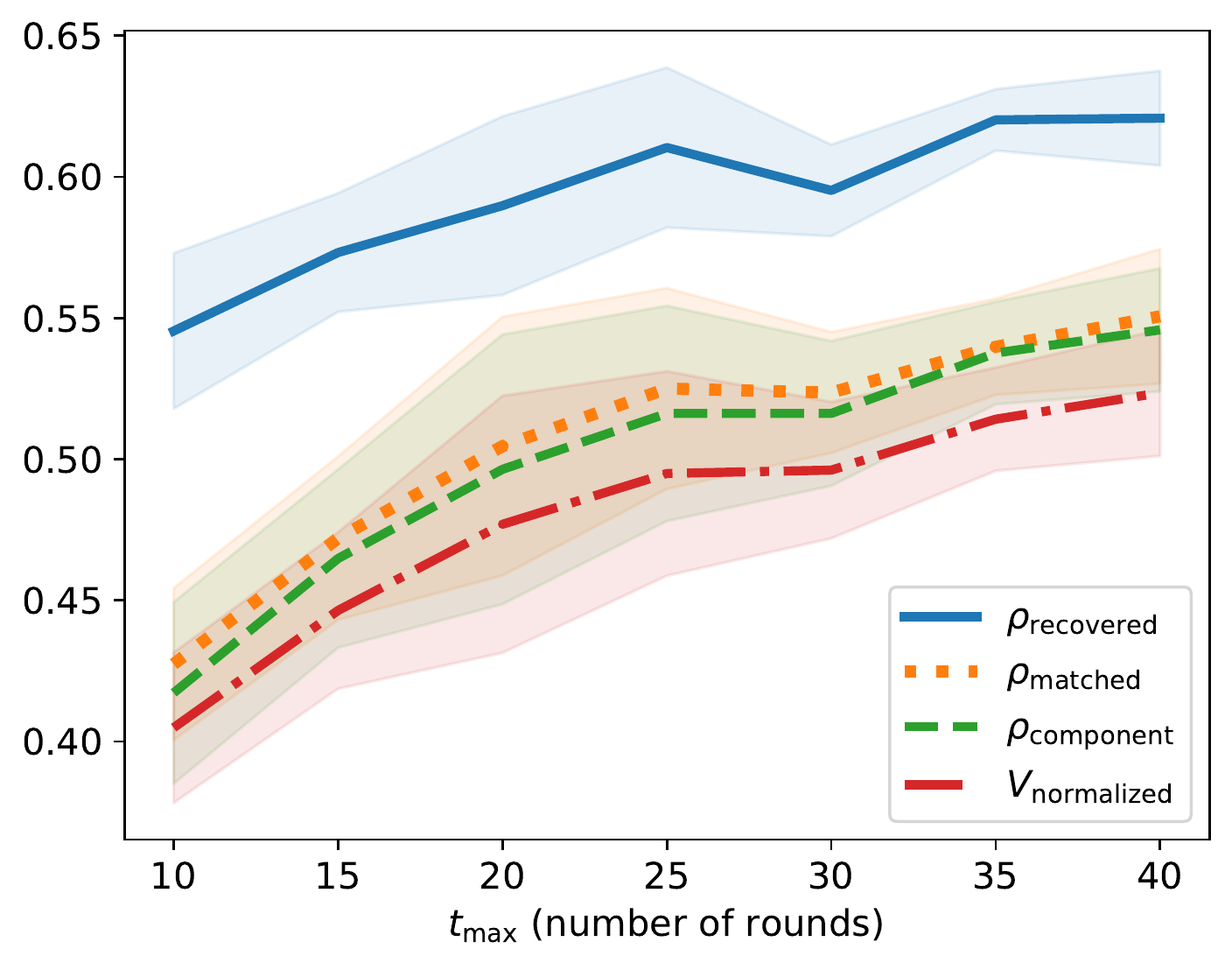}
            \caption{}
      \label{fig:effect_rounds}
   \end{subfigure}
   \caption{Effects of batch size, number of hidden neurons, learning rate and number of rounds on the performance of \att\ on FashionMNIST. Hyper-parameters used are details in~\Cref{tab:hp}.}
   \label{fig:effect_hp_1}
\end{figure}

\begin{figure}[h]
   \centering   
      \begin{subfigure}[b]{0.4\textwidth}
      \includegraphics[width=\textwidth]{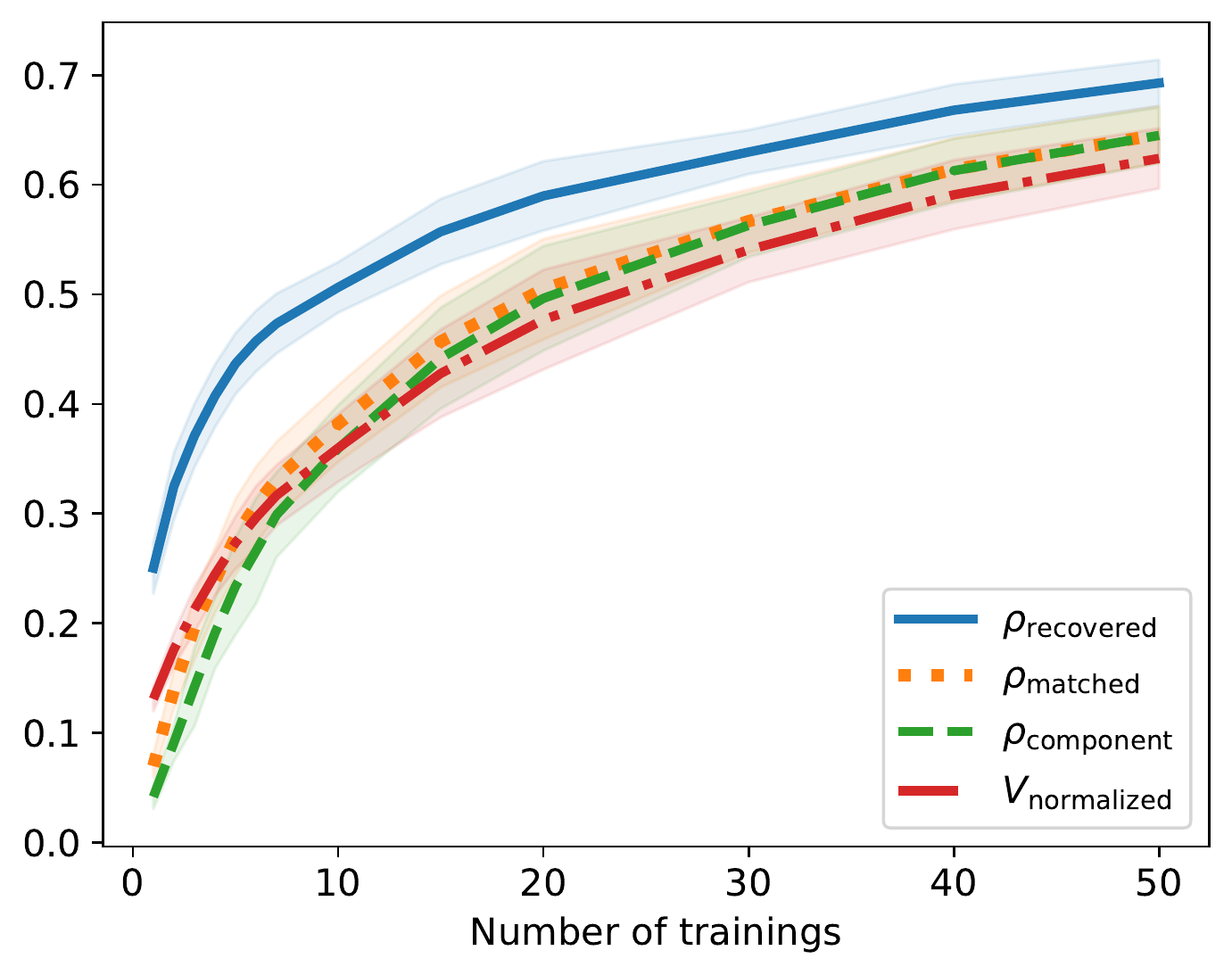}
            \caption{}
      \label{fig:effect_trainings}
   \end{subfigure}
         \begin{subfigure}[b]{0.4\textwidth}
      \includegraphics[width=\textwidth]{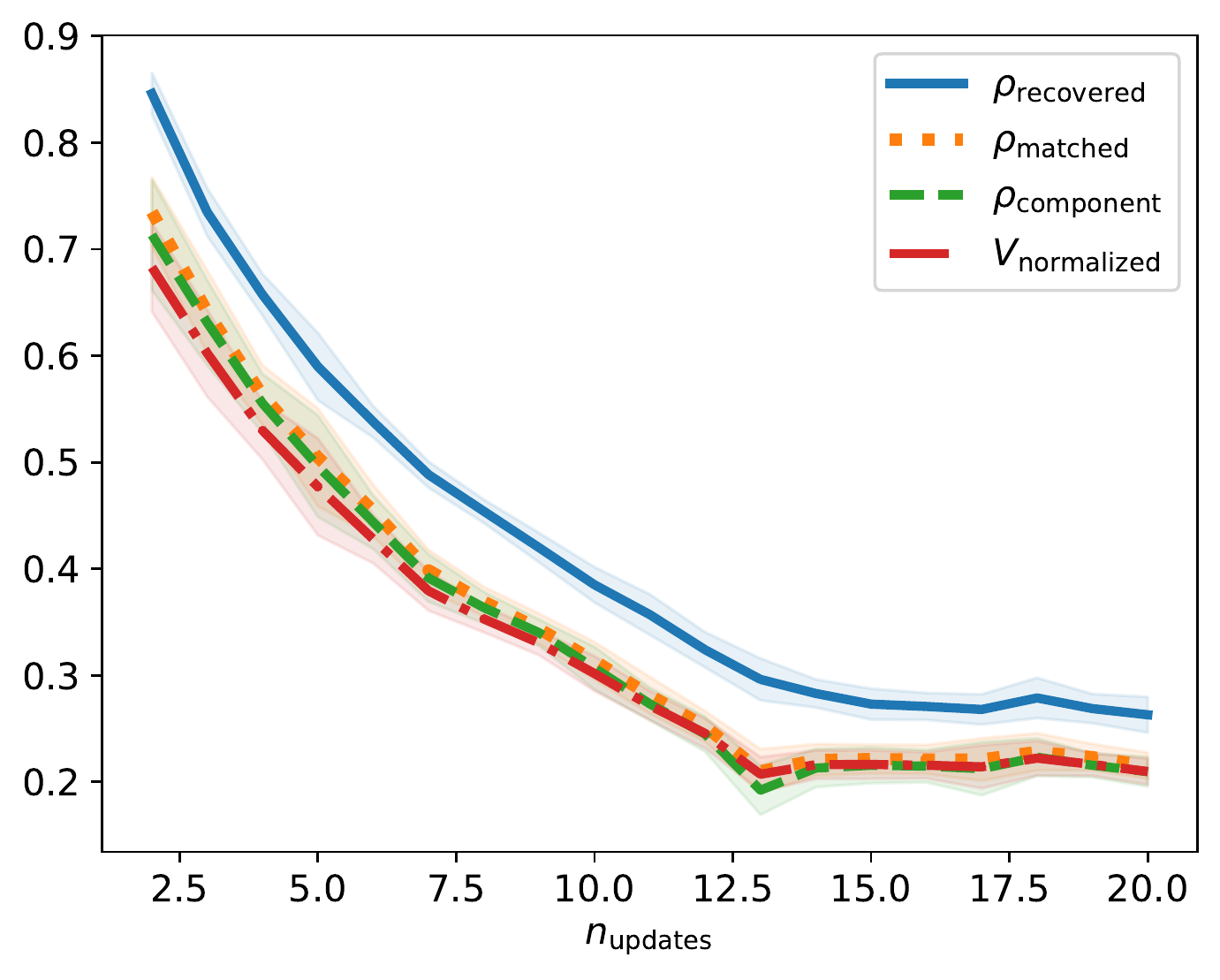}
            \caption{}
      \label{fig:effect_updates}
   \end{subfigure}
            \begin{subfigure}[b]{0.4\textwidth}
      \includegraphics[width=\textwidth]{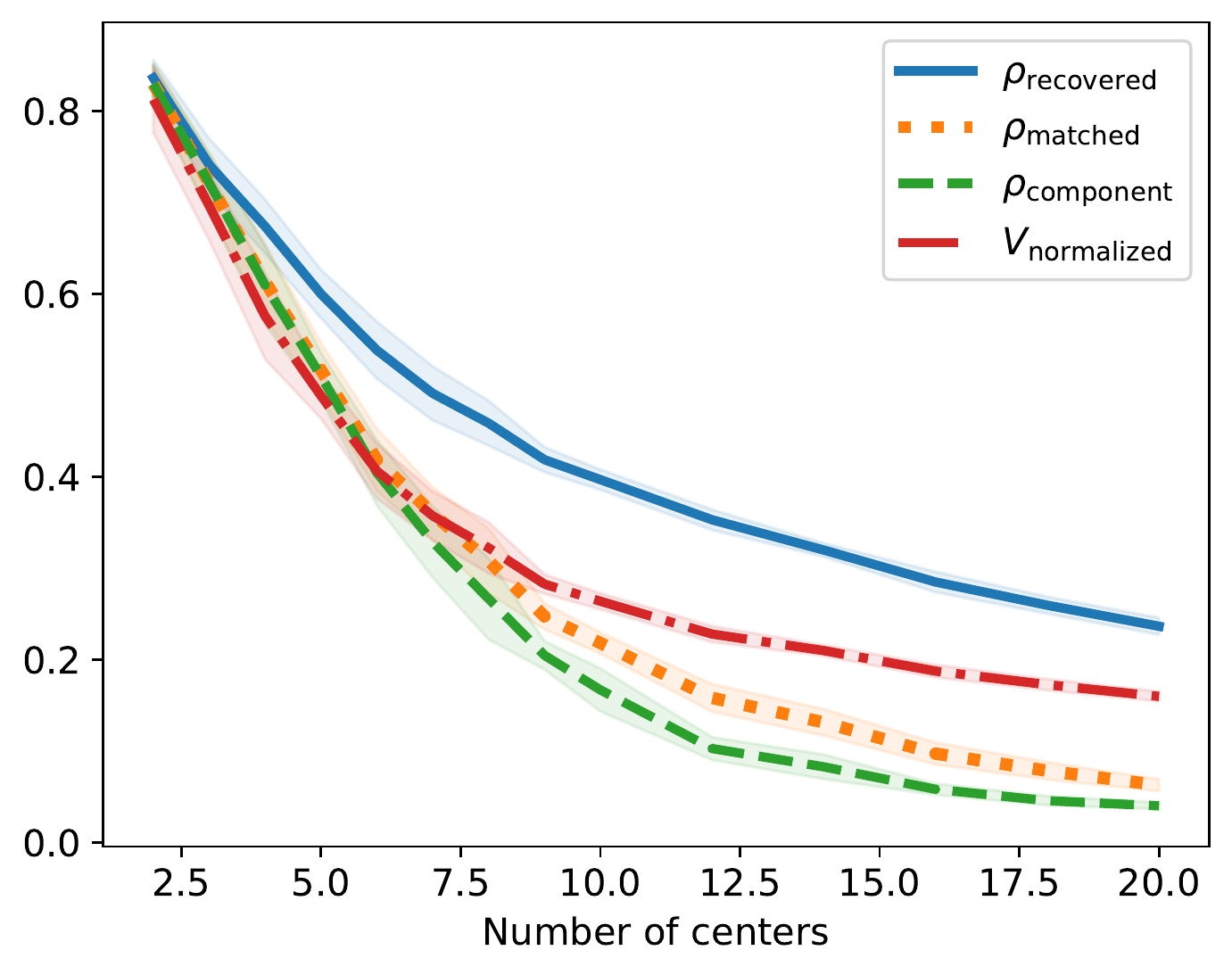}
            \caption{}
      \label{fig:effect_centers}
   \end{subfigure}
            \begin{subfigure}[b]{0.4\textwidth}
      \includegraphics[width=\textwidth]{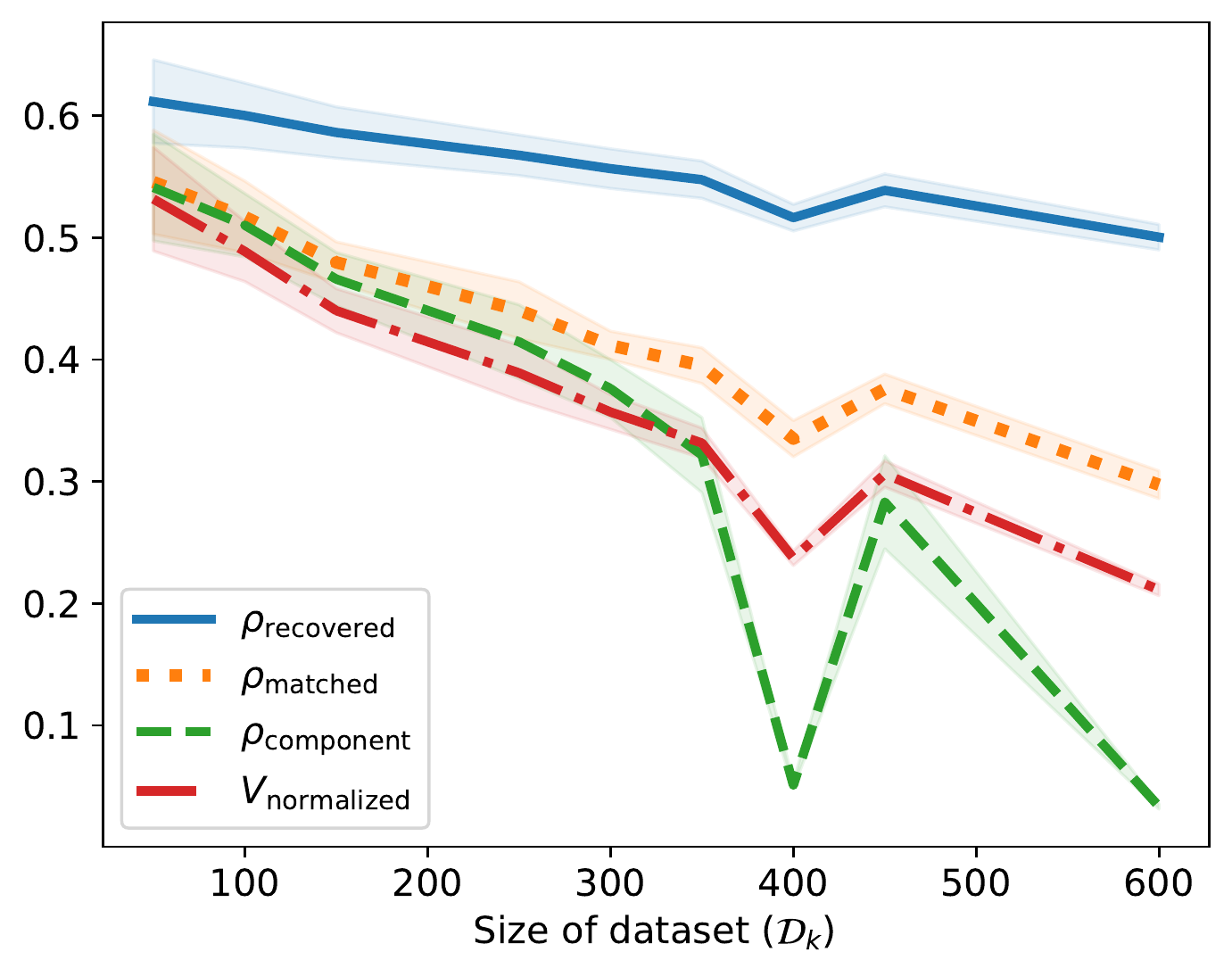}
            \caption{}
      \label{fig:effect_num_datasamples}
         \end{subfigure}
   \caption{Effects of the number of trainings, number of local updates, number of centers and dataset size on the performance of \att\ on FashionMNIST. Hyper-parameters used are details in~\Cref{tab:hp}.}
   \label{fig:effect_hp_2}
\end{figure}

In this section, we give more details as to the effect of different hyper parameters on the strength of \att.~\Cref{fig:effect_hp_1,fig:effect_hp_2} shows the impact of the batch size,
the number of hidden neurons, the learning rate, the number of rounds, the number of trainings and the number of updates per round in the efficiency of \att.

\section{Defending against secure disaggregation}\label{app:defense}

This section is devoted to the defense presented in~\Cref{sec:defense}, that we will name $q$-defense. We first state definitions used in the main text then we provide details and extra-results on the $q$-defense in~\Cref{app:add_def_defense}. 
Then, in~\Cref{sec:def_rel}, we develop and implement another kind of defense, denoted $\beta$-defense, which is similar to the one presented in~\Cref{app:add_def_defense}, but relies on another way of selecting the neurons to censor. 

\subsection{Additional definitions for the $q$-defense}%
\label{app:add_def_defense}

In~\Cref{sec:defense}, given a client $k$, we define the individual client activation set of neuron $h$ at round $t$ as
\begin{equation}
\label{eq:df_activation_client}
\act^h_{t,k}= \bigcup_{i=0}^{\nupdates- 1}{\act^h(\theta_{t,k,i},B_{t,k,i}) }.
\end{equation}
One can write the update of a specific client $k$ as
\begin{equation}
\Delta \Wbar^h_{t,k} \defeq  \Wbar^h_{t,k,\nupdates} - \Wbar_{t-1}^h.
\end{equation}
Exactly as is the case for the round-level activation set, the update of a specific client is a linear combination of the elements in its activation set. 

Fix a client $k$, a neuron $h$, and $t \in \{1,...,\tmax\}$. Let $x_1,...,x_{N}$ denote the activation points in $\act^h_{t,k}$. 
There exists non zero coefficients $(\lambda_r) \in \R\setminus{\{0\}}^{N}$ such that 
\begin{align}\label{eq:client_activations}
\Delta W^h_{t,k} &= \sum_{r=1}^{N}{ \lambda_r x_r}\\
\Delta b^h_t &= \sum_{r=1}^{N}{\lambda_r}
\end{align}
The $q$-defense, introduced in~\Cref{sec:defense}, consists in keeping track of the individual client activation set of all neurons, and censoring the update for neurons whose individual client activation set is non-empty and of size inferior to $q$. 

Considering a client $k$, let fix $q=1$ and $\tilde h$ be such a neuron, i.e:
\begin{equation}
   \# \act^{\tilde h}_{t,k} = 1.
   \end{equation}
For this neuron, there exists a data sample such that $\{x\} = \act^{\tilde h}_{t,k}$, which can be recovered by the first step of the attack, as from~\Cref{eq:activation_reconstruction}:
\begin{subequations}%
\label{eq:activation_reconstruction_aux_alone}
\begin{align}
\Delta W^{\tilde h}_{t,k} &= \lambda x \\
\Delta b^{\tilde h}_{t,k} &= \lambda \\
\lambda &\neq 0
\end{align}
\end{subequations}
To avoid this situation, the defense relies on \textit{censoring} this neuron, which means to set to zero the update sent to the server for $\tilde h$. We define an integer threshold $q$, and censor neurons whose individual client activation sets are smaller than $q$. The resulting algorithm is depicted in~\Cref{alg:localupdate_abs_def}. 
This defense therefore requires an active role of the client.

\begin{algorithm}[H]
   \caption{LocalUpdateDefended // Executed on server k}
         \label{alg:localupdate_abs_def}
\begin{algorithmic}
   \REQUIRE initial model $\theta_{t-1}$, local dataset $\dd_k$, batch size $b$, learning rate $\eta$
   \STATE $\theta_{t, k, i=0} \leftarrow \theta_{t-1}$
   \FOR{$i=0$ {\bfseries to} $\nupdates-1$}
          \STATE $ B_{t,k,i} \leftarrow$ batch of size $b$ from $\dd_k$
         \STATE $\theta_{t, k, i+1} \leftarrow \theta_{t, k, i} - \eta \nabla_\theta \LL(\theta_{t,k,i},B_{t,k,i})$
   \ENDFOR
   \textcolor{red}{
   \FOR{$h=0$ {\bfseries to} $d$}
      \IF{$\#\act^{h}_{t,k} \leq q$}
         \STATE $\Wbar^h_{t,k,\nupdates}  \leftarrow  \Wbar_{t-1}^h$
      \ENDIF
   \ENDFOR
   }
    \ENSURE  $\theta_{t, k,\nupdates}$
\end{algorithmic}
\end{algorithm}

\Cref{fig:defense_acc_and_prun_all_datasets} and~\Cref{tab:def_full} shows the efficiency of the  $q$-defense against the attack. Note that if some samples are still recovered with $q=1$, using $q=4$ leads to a perfect defense (no sample recovered) in our experiments. 
\subsection{Additional results for $q$-defense}

\begin{figure}
   \centering

   \begin{subfigure}[b]{0.4\textwidth}
      \centering
      \includegraphics[width=\textwidth]{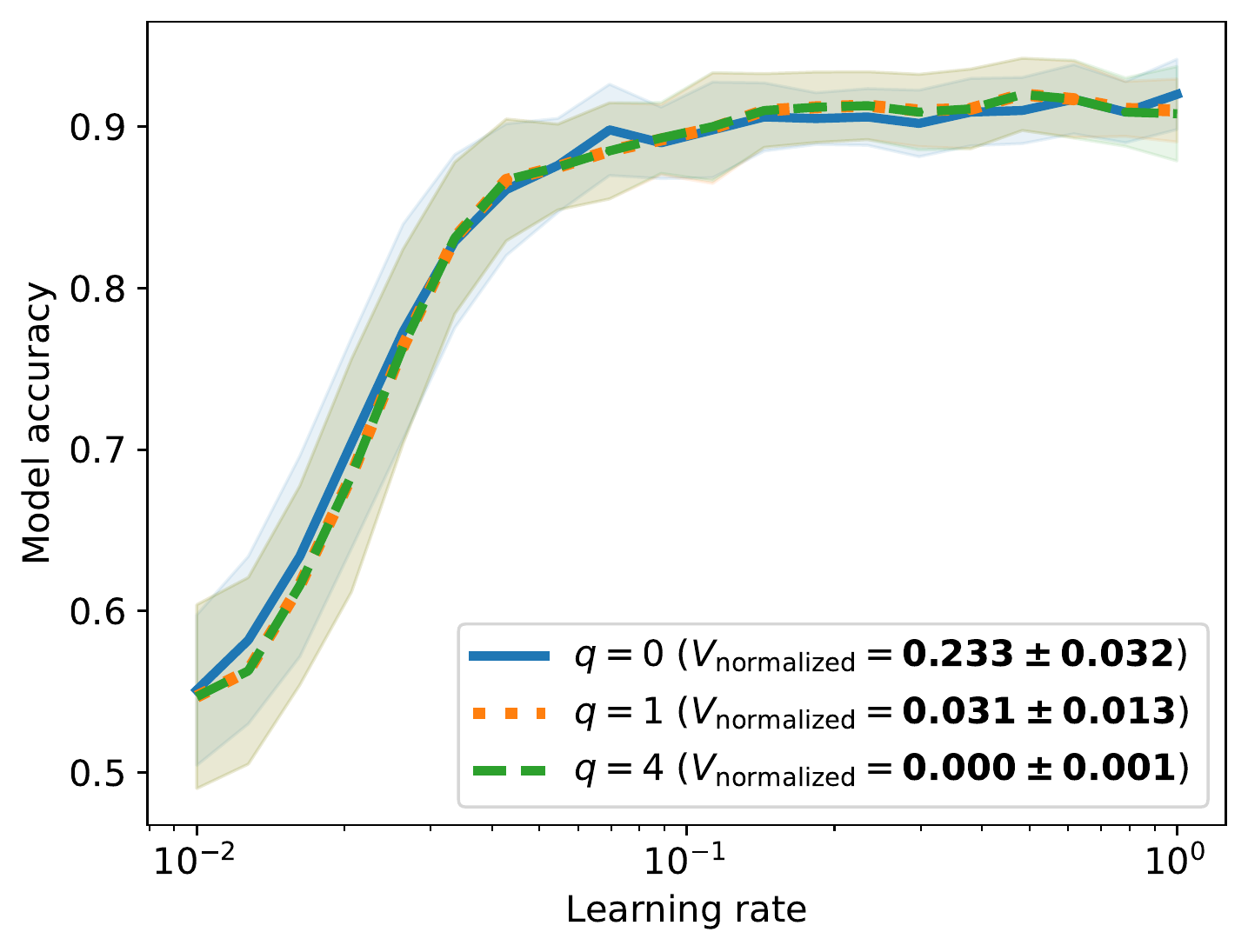}
      \caption{DNA}
      \label{fig:defense_acc_dna}
   \end{subfigure}
   \begin{subfigure}[b]{0.4\textwidth}
      \centering
      \includegraphics[width=\textwidth]{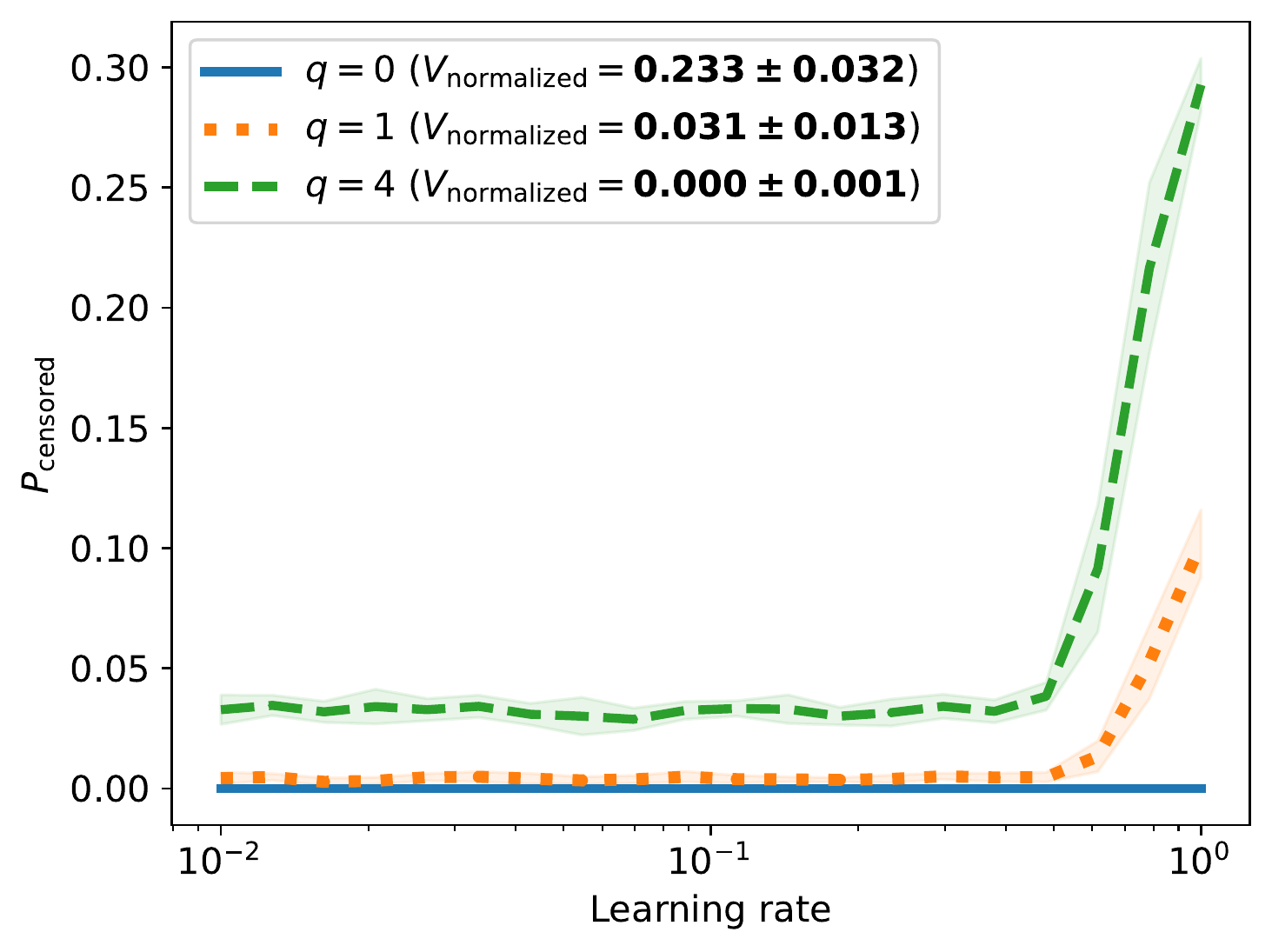}
      \caption{DNA}
      \label{fig:defense_prun_dna}
   \end{subfigure}

   \begin{subfigure}[b]{0.4\textwidth}
      \centering
      \includegraphics[width=\textwidth]{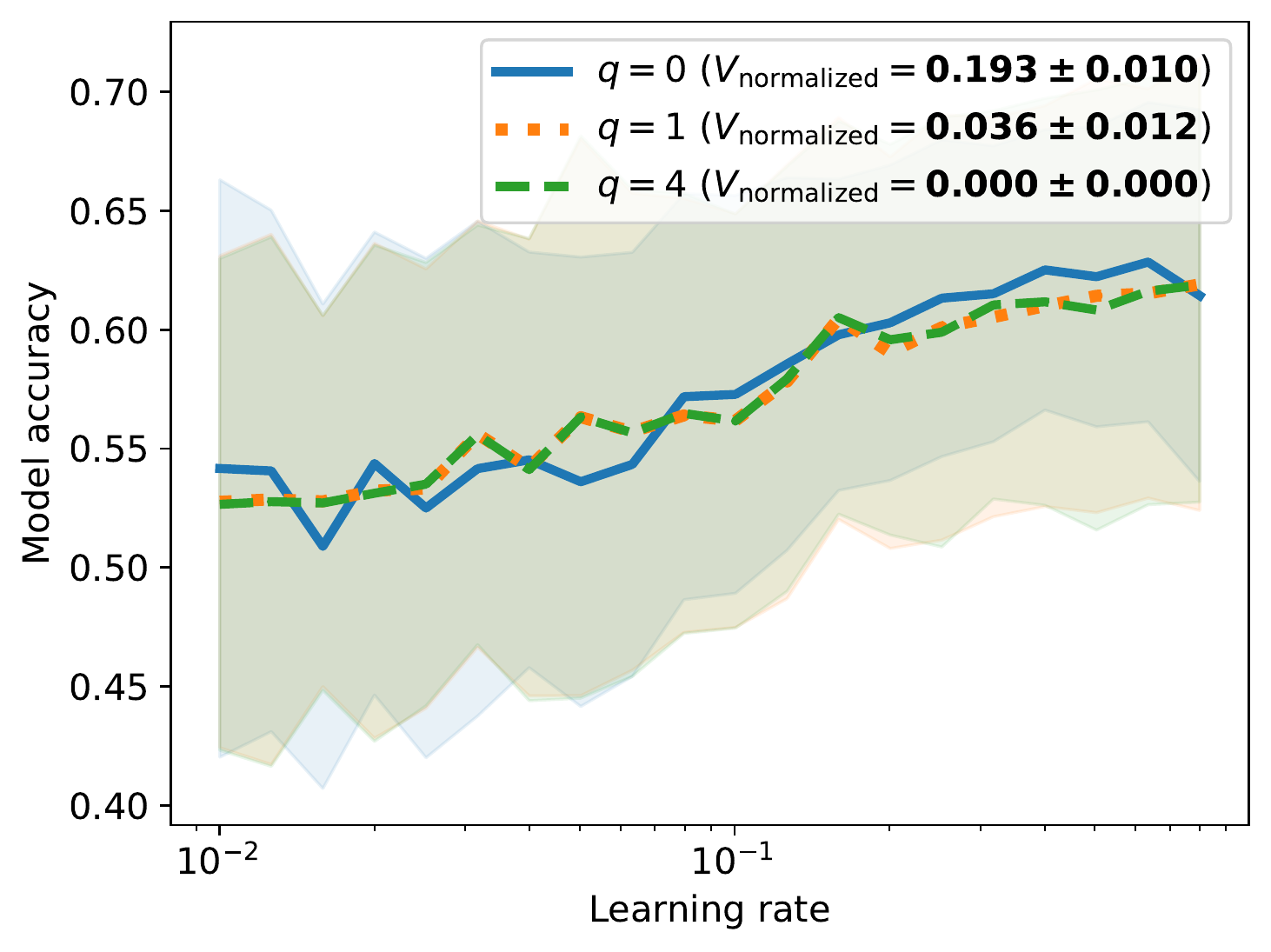}
      \caption{TCGA}
      \label{fig:defense_acc_tcga}
   \end{subfigure}
   \begin{subfigure}[b]{0.4\textwidth}
      \centering
      \includegraphics[width=\textwidth]{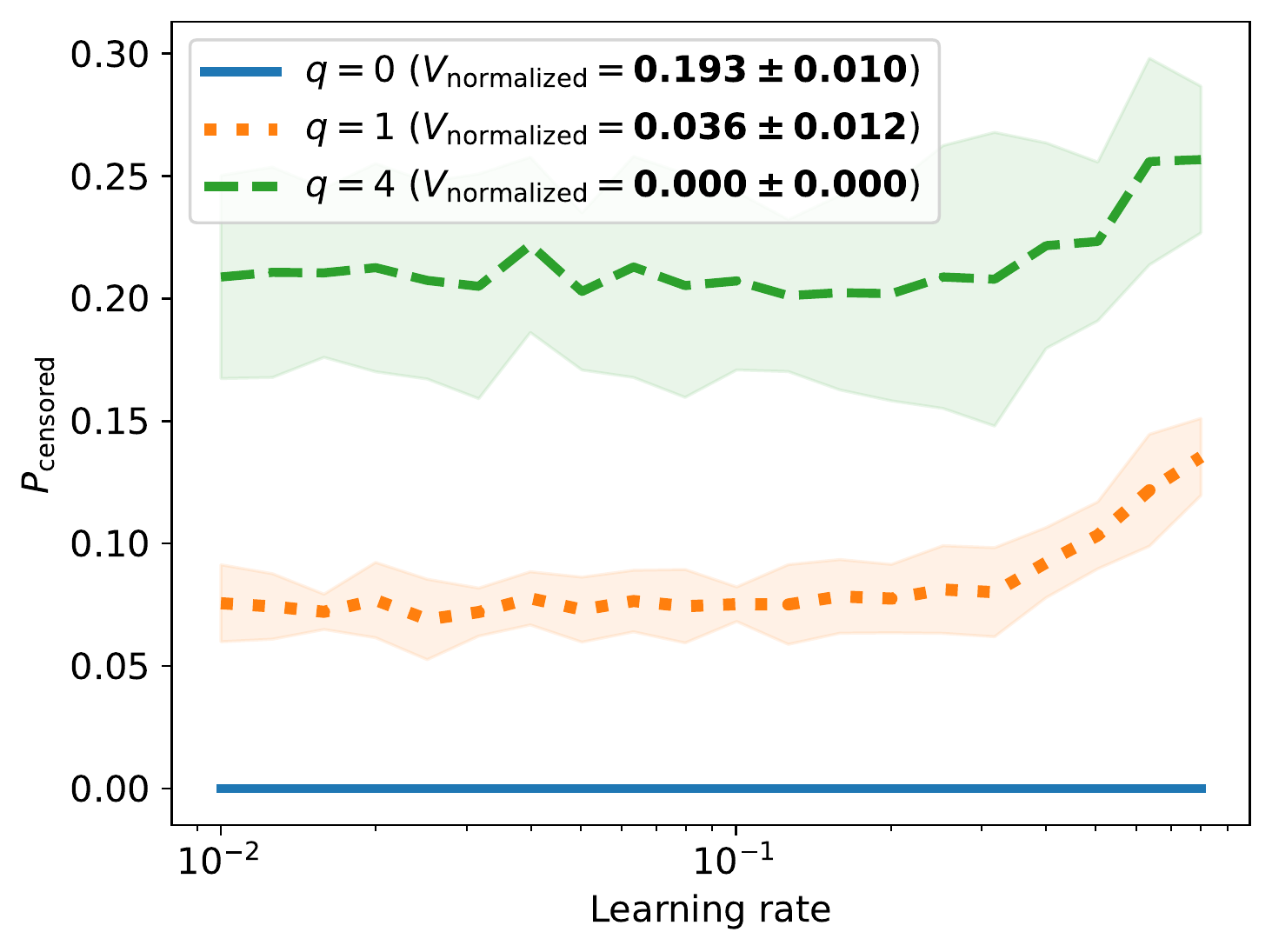}
      \caption{TCGA}
      \label{fig:defense_prun_tcga}
   \end{subfigure}
      
   \caption{Effects of the $q$-defense (presented in~\Cref{sec:defense} and detailed in~\Cref{app:add_def_defense}) on the model accuracy and the number of neurons frozen. $P_\mathrm{censored}$ corresponds to the proportion of neurons censored compared to the total number of neurons.}
         \label{fig:defense_acc_and_prun_all_datasets_2}
\end{figure}

\Cref{tab:def_full} shows the complete results of the $q$-defense on the different datasets, and~\Cref{fig:defense_prun_tcga} shows the proportion of censored neurons by  $q$-defense for different learning rates.
\begin{figure}
   \centering
   \begin{subfigure}[b]{0.4\textwidth}
      \centering
      \includegraphics[width=\textwidth]{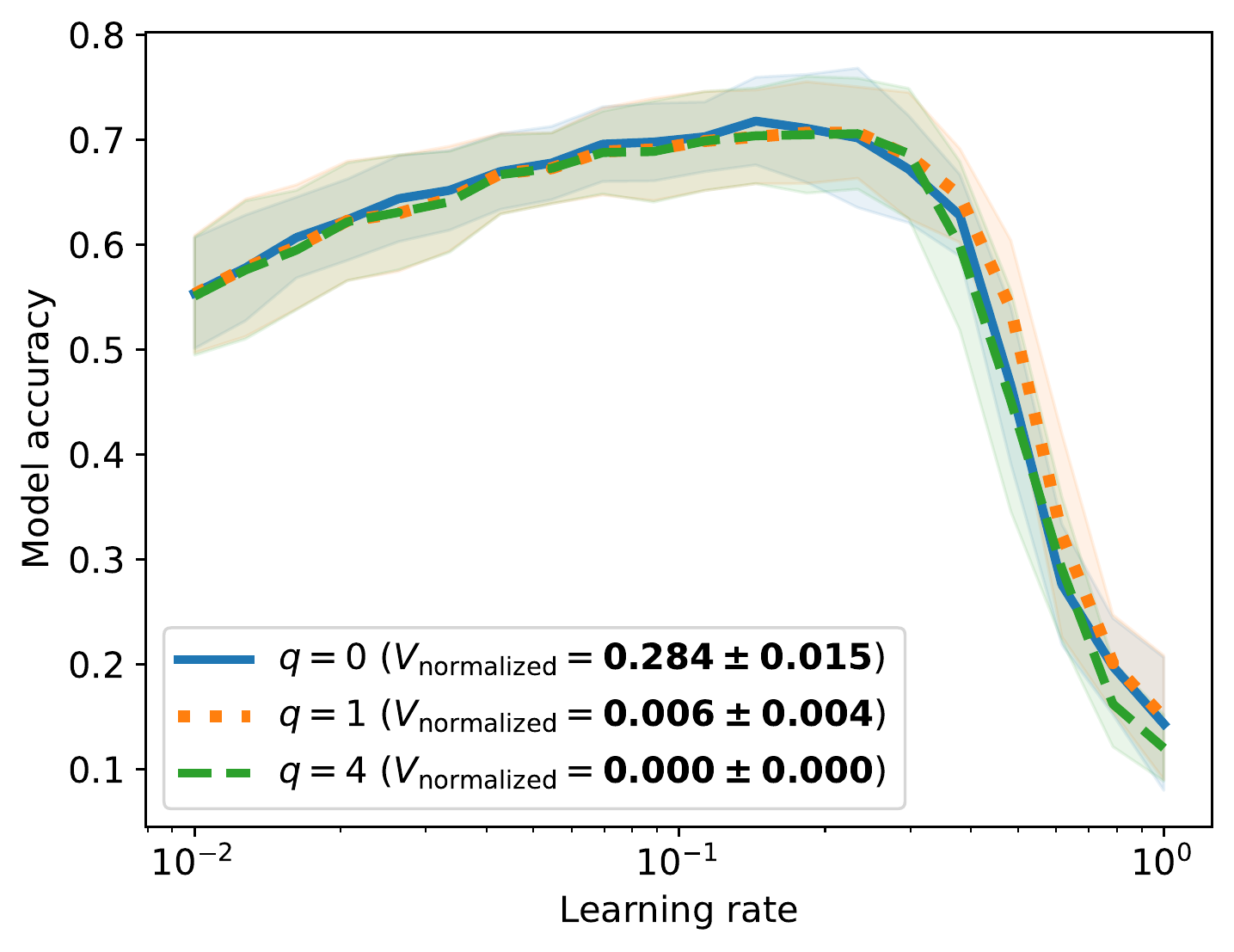}
      \caption{FashionMNIST}
      \label{fig:defense_acc_fmnist}
   \end{subfigure}
   \begin{subfigure}[b]{0.4\textwidth}
      \centering
      \includegraphics[width=\textwidth]{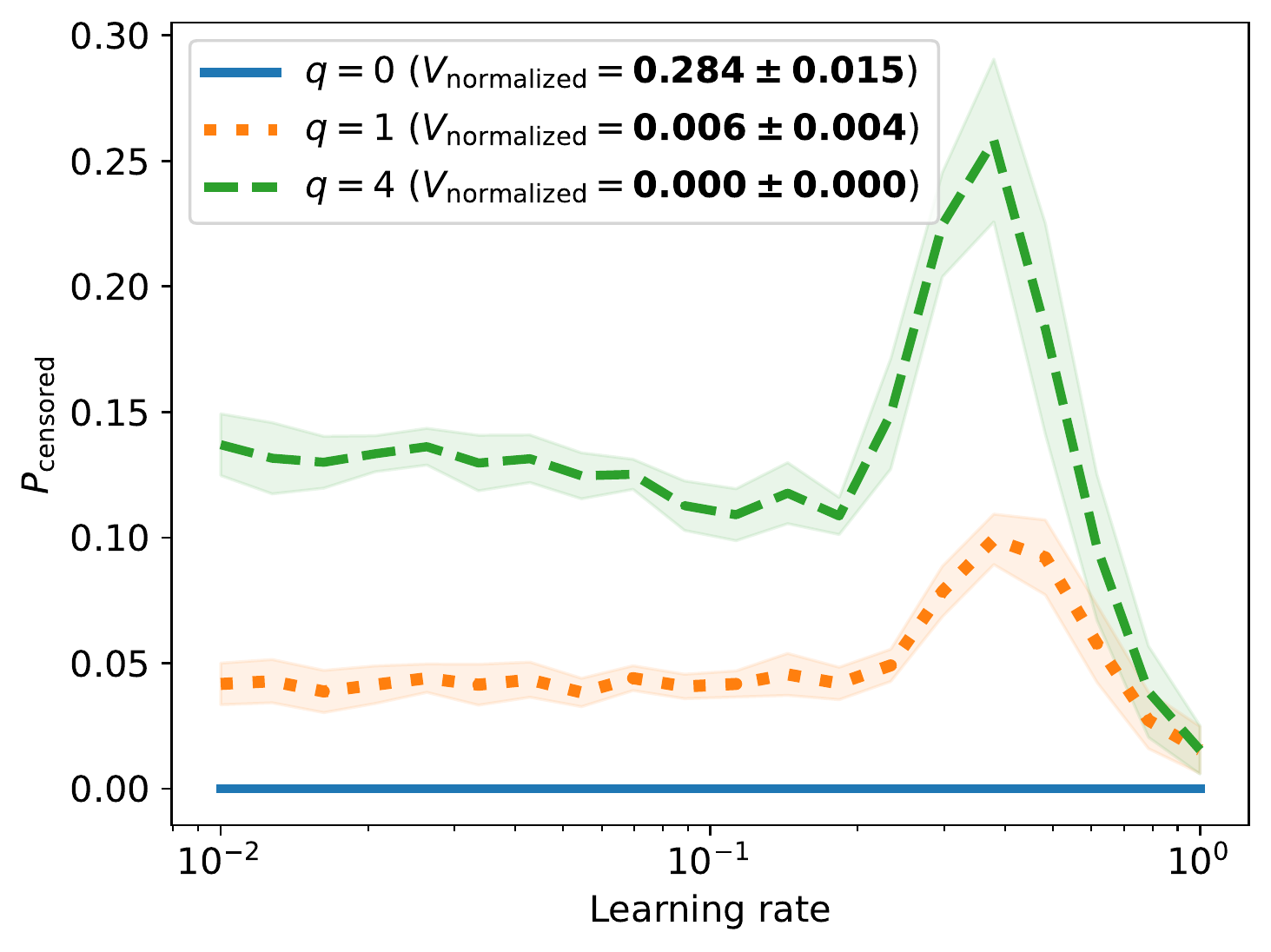}
      \caption{FashionMNIST}
      \label{fig:defense_prun_fmnist}
   \end{subfigure}

   \begin{subfigure}[b]{0.4\textwidth}
      \centering
      \includegraphics[width=\textwidth]{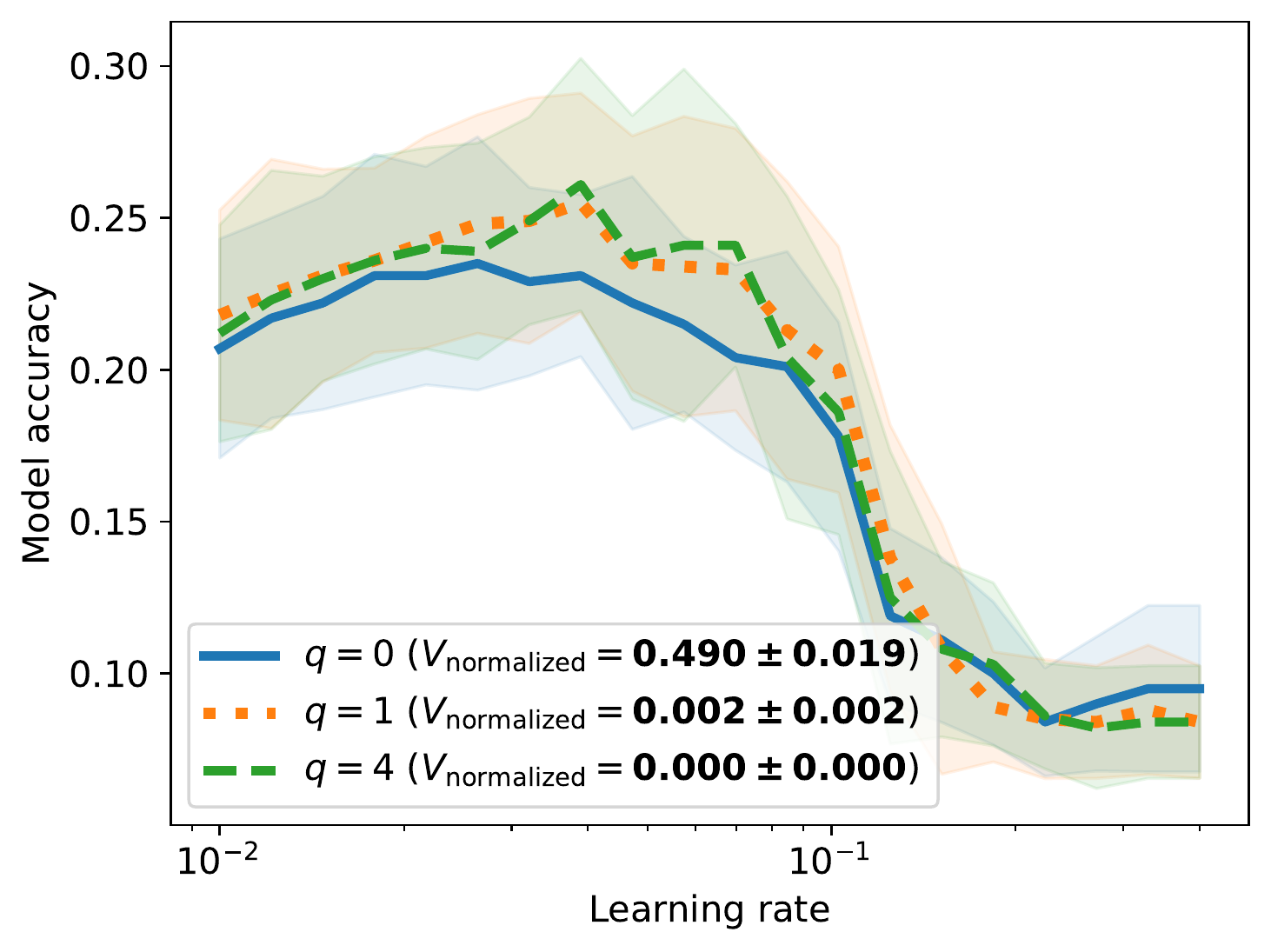}
      \caption{CIFAR10}
      \label{fig:defense_acc_cifar10}
   \end{subfigure}
   \begin{subfigure}[b]{0.4\textwidth}
      \centering
      \includegraphics[width=\textwidth]{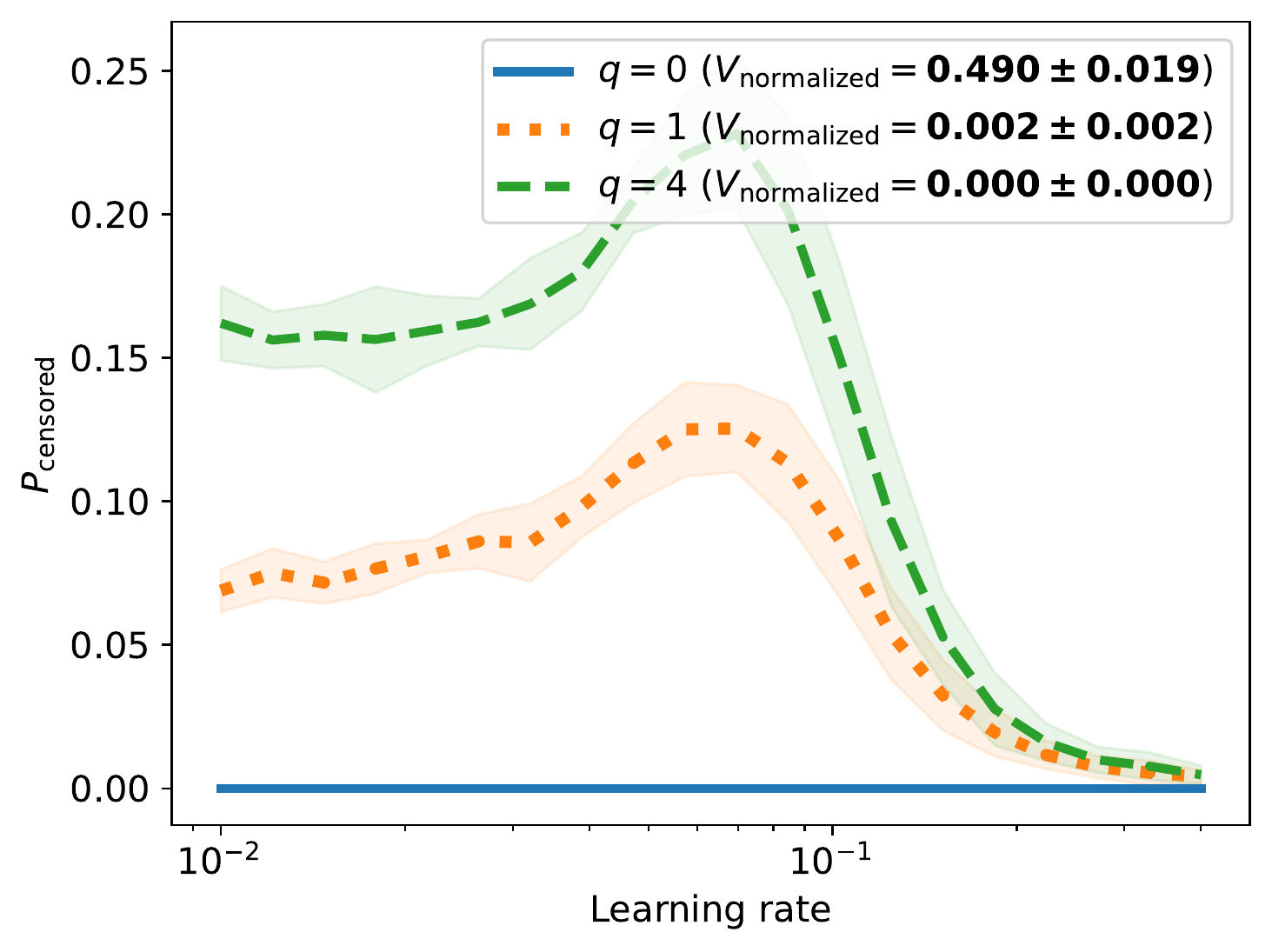}
      \caption{CIFAR10}
      \label{fig:defense_prun_cifar10}
   \end{subfigure}
      
   \caption{Effects of the $q$-defense (presented in~\Cref{sec:defense} and detailed in~\Cref{app:add_def_defense}) on the model accuracy and the number of neurons frozen. $P_\mathrm{censored}$ corresponds to the proportion of neurons censored compared to the total number of neurons.}
         \label{fig:defense_acc_and_prun_all_datasets_1}
\end{figure}

\begin{table}[h]
   \centering\resizebox{\columnwidth}{!}{
   \begin{tabular}{lcccccc}
   \toprule
     Dataset  & $\rrecov \downarrow $   & $\rmatched  \downarrow $   & $\rcomponent \downarrow $   & $\vweighed \downarrow $   & $P_\mathrm{censored}$   & Model Acc. $\uparrow$ \\
   \hline 
   CIFAR10 ($q=0$)      & $0.585 \pm 0.015$        & $0.514 \pm 0.021$      & $0.511 \pm 1.951$       & $0.490 \pm 0.019$         & $0.000 \pm 0.000$     & $0.261 \pm 0.039$ \\
   CIFAR10 ($q=1$)      & $0.002 \pm 0.002$        & $0.000 \pm 0.001$      & $0.000 \pm 0.000$        & $0.002 \pm 0.002$         & $0.098 \pm 0.011$     & $0.272 \pm 0.033$ \\
   CIFAR10 ($q=4$)      & $\boldblue{0.000 \pm 0.000}$        & $\boldblue{0.000 \pm 0.000}$      &$\boldblue{0.000 \pm 0.000}$       & $\boldblue{0.000 \pm 0.000}$         & $0.180 \pm 0.014$     & $\boldsymbol{0.274 \pm 0.040}$ \\
   \midrule
   DNA ($q=0$)          & $0.516 \pm 0.080$        & $0.031 \pm 0.018$      & $0.022 \pm 0.296$        & $0.233 \pm 0.032$         & $0.000 \pm 0.000$     & $0.929 \pm 0.017$ \\
   DNA ($q=1$)          & $0.035 \pm 0.015$        & $0.000 \pm 0.000$      & $0.000 \pm 0.000$        & $0.031 \pm 0.013$         & $0.005 \pm 0.002$     & $0.929 \pm 0.019$ \\
   DNA ($q=4$)          & $\boldblue{0.000 \pm 0.001}$       & $\boldblue{0.000 \pm 0.000}$     & $\boldblue{0.000 \pm 0.000}$       & $\boldblue{0.000 \pm 0.001}$        & $0.038 \pm 0.006$     & $\boldsymbol{0.932 \pm 0.020}$ \\
   \midrule
   FashionMNIST ($q=0$) & $0.476 \pm 0.027$        & $0.285 \pm 0.022$      & $0.230 \pm 0.015$        & $0.284 \pm 0.015$         & $0.000 \pm 0.000$       & $\boldsymbol{0.732 \pm 0.053}$ \\
   FashionMNIST ($q=1$) & $0.009 \pm 0.004$        & $0.001 \pm 0.003$      & $0.000 \pm 0.000$        & $0.006 \pm 0.004$         & $0.042 \pm 0.006$       & $0.729 \pm 0.043$ \\
   FashionMNIST ($q=4$) & $\boldblue{0.000 \pm 0.000}$        & $\boldblue{0.000 \pm 0.000}$      & $\boldblue{0.000 \pm 0.000}$        & $\boldblue{0.000 \pm 0.000}$         & $0.149 \pm 0.022$       & $0.731 \pm 0.046$ \\
   \midrule
   TCGA ($q=0$)         & $0.279 \pm 0.016$        & $0.199 \pm 0.014$      & $0.175 \pm 1.349$       & $0.193 \pm 0.010$         & $0.000 \pm 0.000$     & $\boldsymbol{0.660 \pm 0.065}$ \\
   TCGA ($q=1$)         & $0.051 \pm 0.016$        & $0.024 \pm 0.017$      & $3.040 \pm 1.158$        & $0.036 \pm 0.012$         & $0.135 \pm 0.016$     & $0.658 \pm 0.071$ \\
   TCGA ($q=4$)         & $\boldblue{0.000 \pm 0.000}$        & $\boldblue{0.000 \pm 0.000}$      &$\boldblue{0.000 \pm 0.000}$        & $\boldblue{0.000 \pm 0.000}$         & $0.257 \pm 0.030$     & $0.658 \pm 0.071$ \\

   \bottomrule
   \end{tabular}}
   \caption{Influence of the $q$-defense (presented in~\Cref{sec:defense} and detailed in~\Cref{app:add_def_defense}) on both the attack efficiency and the training performance. 
   The attack is performed on a simulation of a grid search, with 20 training with different learning rates. 
   The model accuracy reported is the best one for all the learning rates tested. This setting highlights 
   the defense success to prevent the attack for a range of learning rates, with no-significant loss of model accuracy. }
   \label{tab:def_full}
   \end{table}

Both show the efficiency of the $q$-defense against \att. Note that if some samples are still recovered with $q=1$, using $q=4$ leads to perfect defense (no sample recovered) in our experiments. In practice, samples can be recovered if $\act^{\tilde h}_{t,k}$ is not a singleton, in the case where one of the $\lambda_r$ in~\Cref{eq:client_activations} is an order of magnitude larger than the others in absolute value.

\subsection{Another possible defense: the $\beta$-defense}%
 \label{sec:def_rel}
   We introduce in this section another defense, named "$\beta$-defense".
   Increasing the value of $q$ is an efficient way to avoid sample recovery. However, it might censor more neurons than needed. Indeed, as mentioned above, if $\#\act^{\tilde h}_{t,k}>1$ for a given neuron $\tilde h$, a sample $x_r$ can be recovered if its associated $\lambda_r$ in~\Cref{eq:round_activations} is larger than the other $\lambda_j$ by an order of magnitude. Note that for each neuron $h$, for each sample $x_r$ involved in the local training, the corresponding $\lambda^h_r$
   can be computed by the client. Following the computation of~\Cref{proof_t}, one can derive
   \[\Delta W^{h}_{t,k} = -\eta \sum_{i=0}^{\nupdates-1}{\sum_{j=0}^{b} \lambda(\theta_{t,k,i}; x_j^{B^{i}_{k}}, y_j^{B^{i}_{k}}) x_j}\]
\[\Delta b^{h}_{t,k} = -\eta \sum_{i=0}^{\nupdates-1}{\sum_{j=0}^{b} \lambda(\theta_{t,k,i}; x_j^{B^{i}_{k}}, y_j^{B^{i}_{k}})}\]
with $\lambda(\theta;x,y)  =\frac{\partial \LL(\theta;x,y)}{\partial z^h}  \mathbf{1}_{W^h x+b^h> 0}$. These last quantities are tractable by the client during the local update. We introduce a threshold $\beta$ and we censor all neurons for which there exists $\tilde x \in \bigcup_{i=0}^{\nupdates- 1}{\act^h(\theta_{t,k,i},B_{t,k,i}) }$, and its associated $\tilde \lambda$ such that:

\begin{equation}
   \frac{\vert \tilde \lambda \vert }{\sum_{i=0}^{\nupdates-1}{\sum_{j=0}^{b} \vert \lambda(\theta_{t,k,i}; x_j^{B^{i}_{k}}, y_j^{B^{i}_{k}})\vert}} \geq \beta 
\end{equation} 

In plain english, if the linear weight of a sample contributes to the update more than a fraction $\beta$, we censor the neuron. We name this other defense $\beta$-defense, different to the $q$-defense in the way it selects the neurons to censor.
The results exhibited in~\Cref{fig:defense_acc_and_prun_all_datasets_1,fig:defense_acc_and_prun_all_datasets_2,tab:def_rel_full} highlights that we are able while censoring less neurons using the $\beta$-defense than when using the $q$-defense with $q=4$, we successfully defend ourselves against the attack. Note that in these results the $q$-defense is never used, as the $\beta$-defense comes as a replacement of the $q$-defense.

\begin{figure}
   \centering
   \begin{subfigure}[b]{0.4\textwidth}
      \centering
      \includegraphics[width=\textwidth]{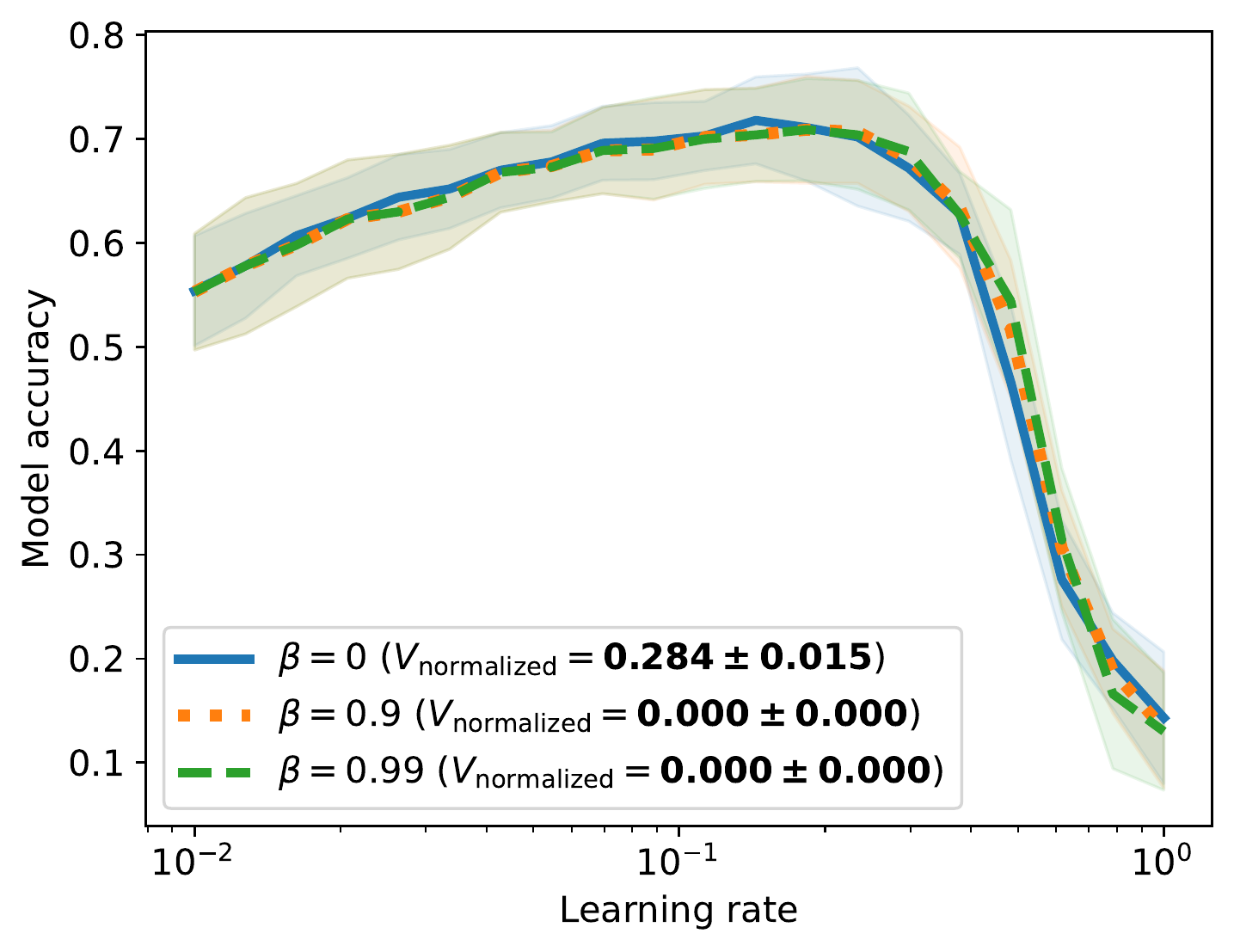}
      \caption{FashionMNIST}
      \label{fig:rel_defense_acc_fmnist}
   \end{subfigure}
   \begin{subfigure}[b]{0.4\textwidth}
      \centering
      \includegraphics[width=\textwidth]{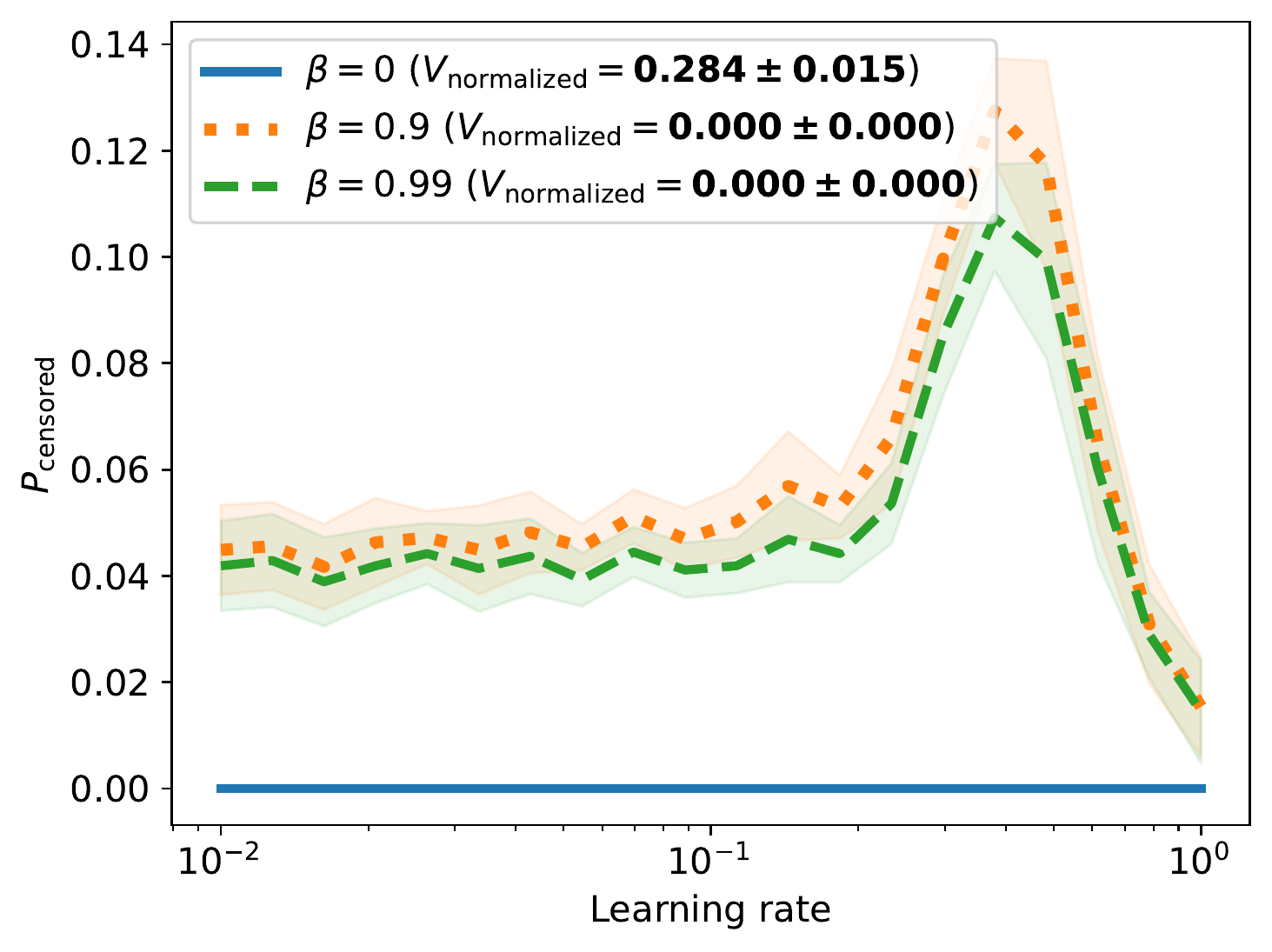}
      \caption{FashionMNIST}
      \label{fig:rel_defense_prun_fmnist}
   \end{subfigure}

   \begin{subfigure}[b]{0.4\textwidth}
      \centering
      \includegraphics[width=\textwidth]{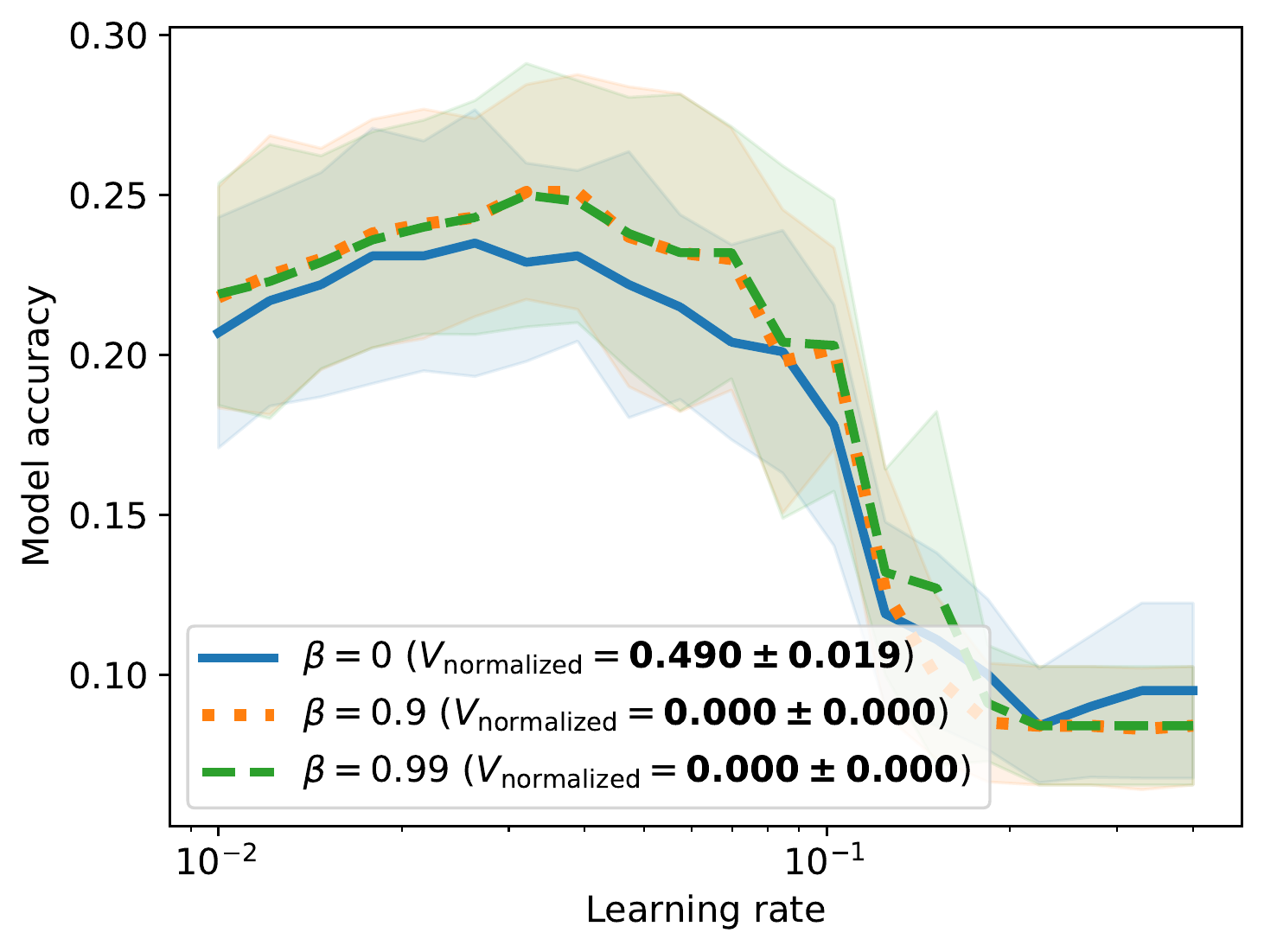}
      \caption{CIFAR10}
      \label{fig:rel_defense_acc_cifar10}
   \end{subfigure}
   \begin{subfigure}[b]{0.4\textwidth}
      \centering
      \includegraphics[width=\textwidth]{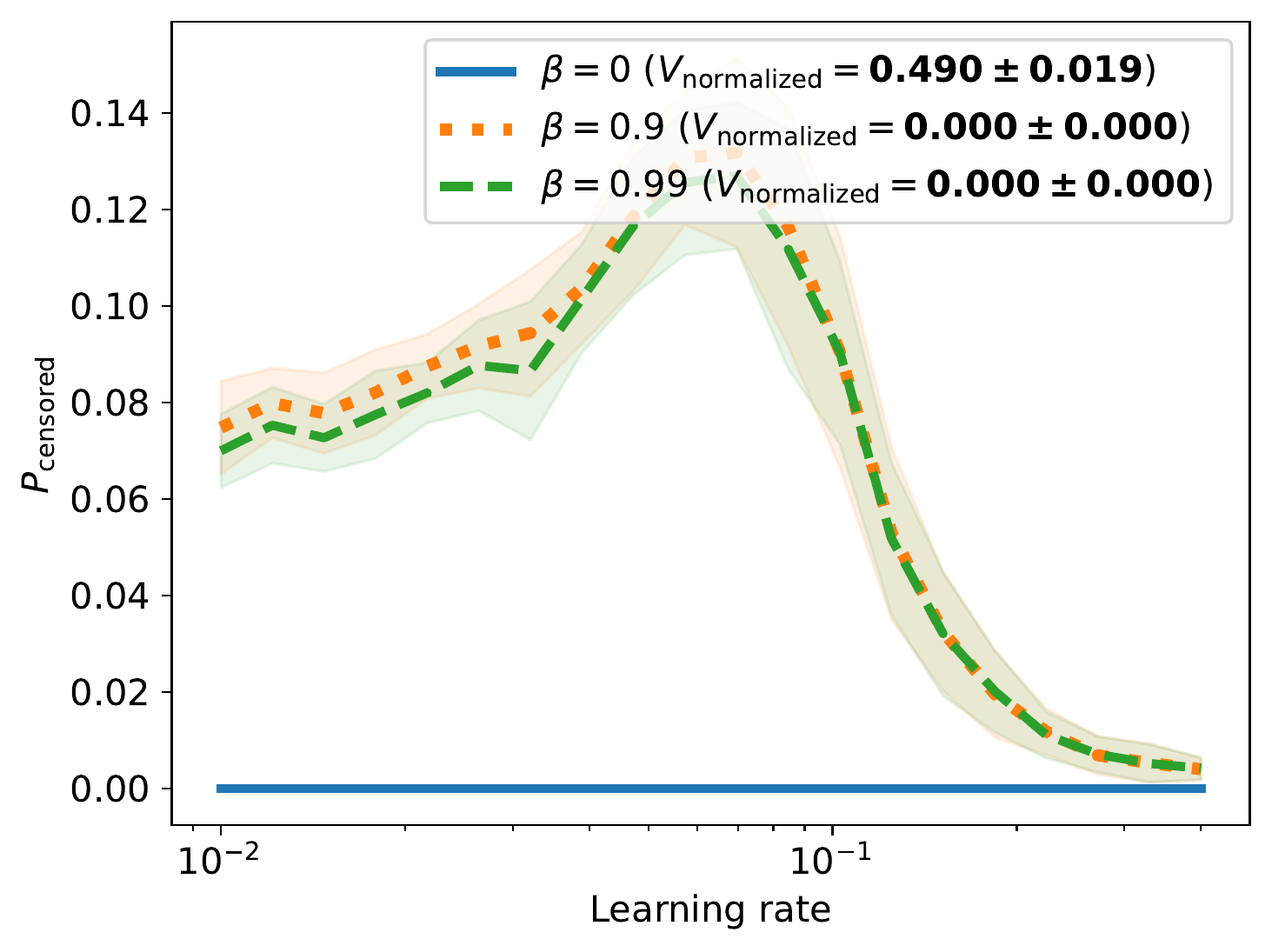}
      \caption{CIFAR10}
      \label{fig:rel_defense_prun_cifar10}
   \end{subfigure}

   \caption{Effects of the  $\beta$-defense on FashionMNIST and CIFAR (presented in~\Cref{sec:def_rel}) on the model accuracy and the number of neurons frozen. $P_\mathrm{censored}$ corresponds to the proportion of neurons censored compared to the total number of neurons.$\beta=0$ means that no defense is applied, neither $\beta$-defense nor $q$-defense.}
         \label{fig:rel_defense_acc_and_prun_all_datasets_1}
\end{figure}

\begin{figure}
   \centering

   \begin{subfigure}[b]{0.4\textwidth}
      \centering
      \includegraphics[width=\textwidth]{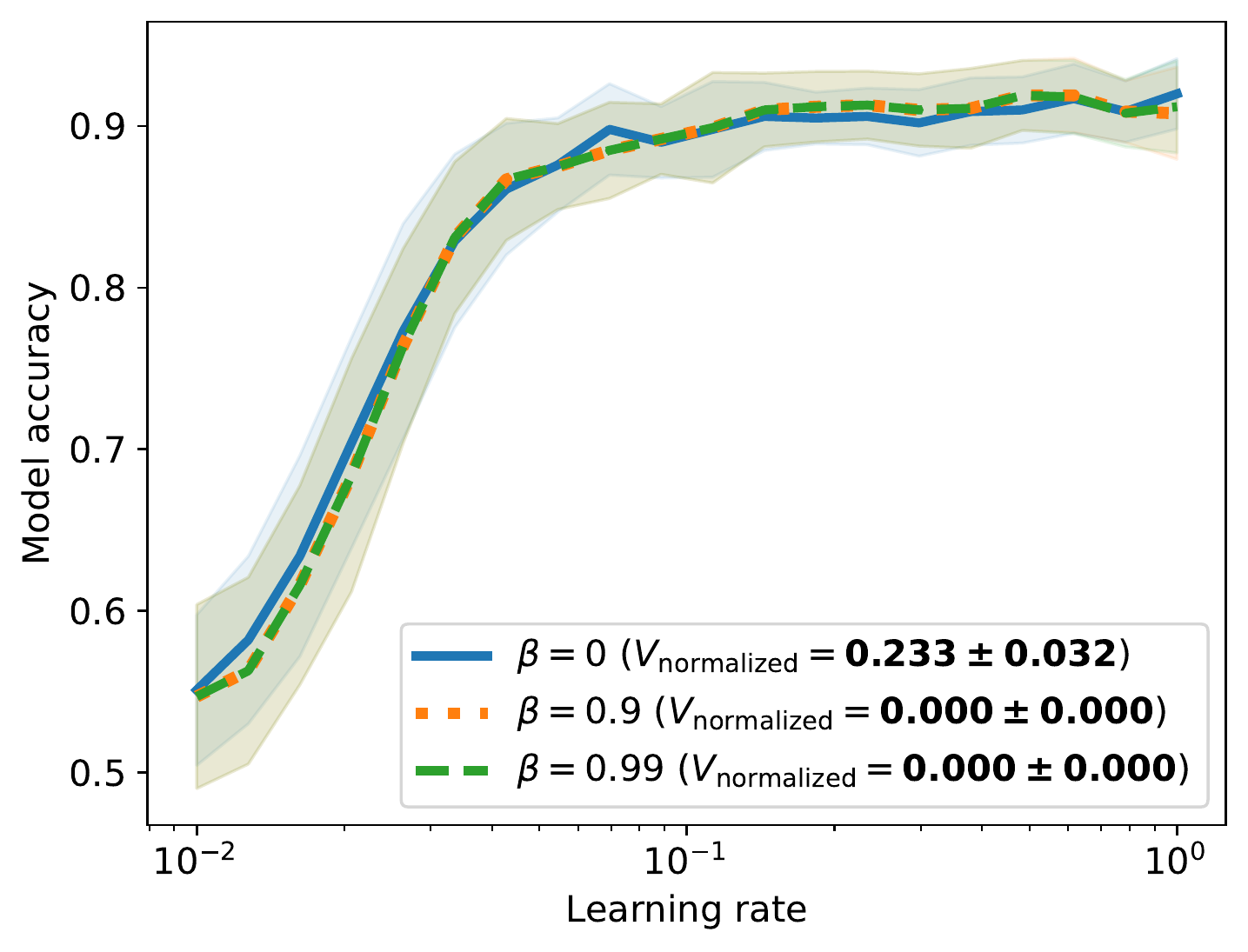}
      \caption{DNA}
      \label{fig:rel_defense_acc_dna}
   \end{subfigure}
   \begin{subfigure}[b]{0.4\textwidth}
      \centering
      \includegraphics[width=\textwidth]{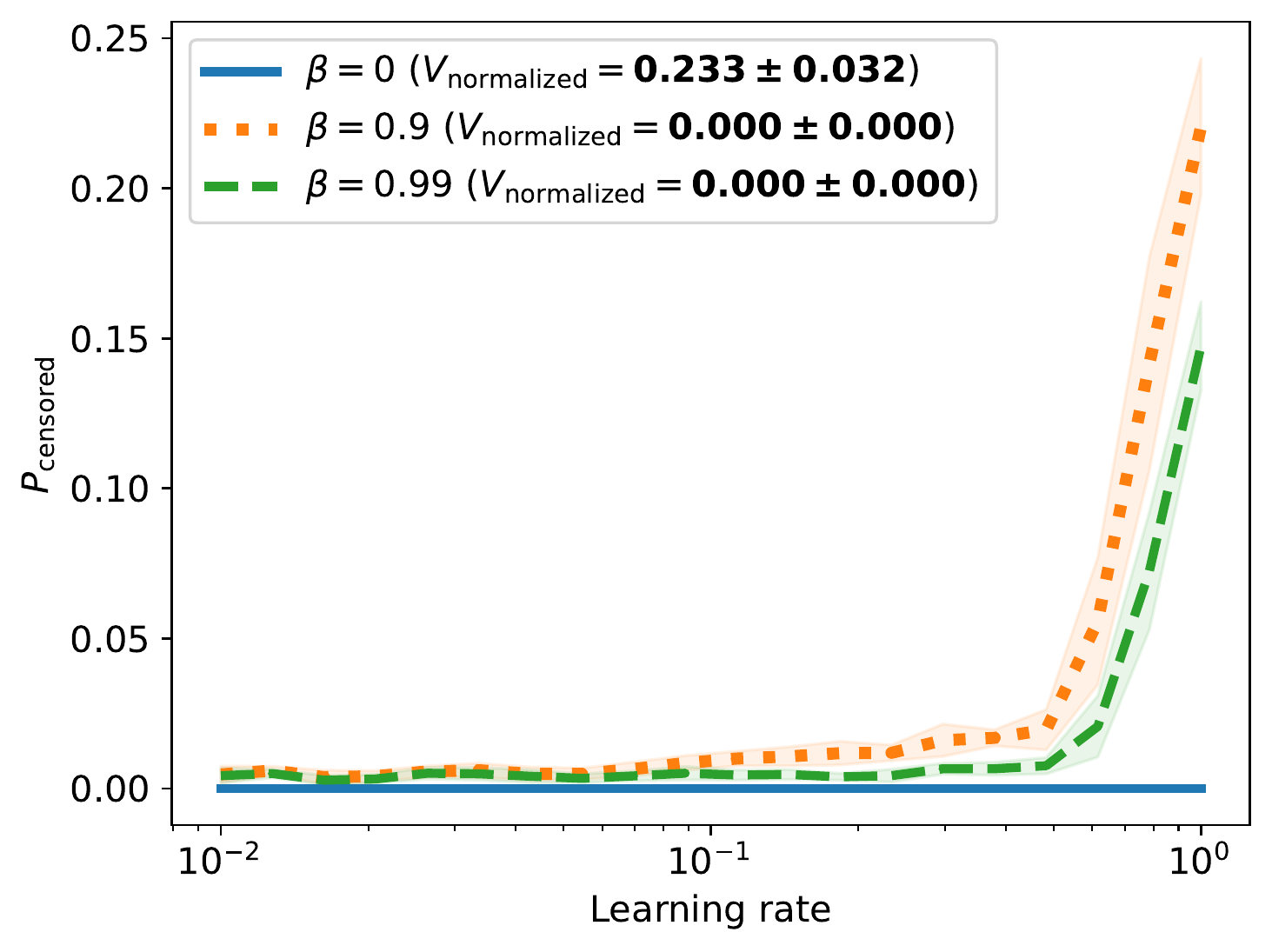}
      \caption{DNA}        
      \label{fig:rel_defense_prun_dna}
   \end{subfigure}

   \begin{subfigure}[b]{0.4\textwidth}
      \centering
      \includegraphics[width=\textwidth]{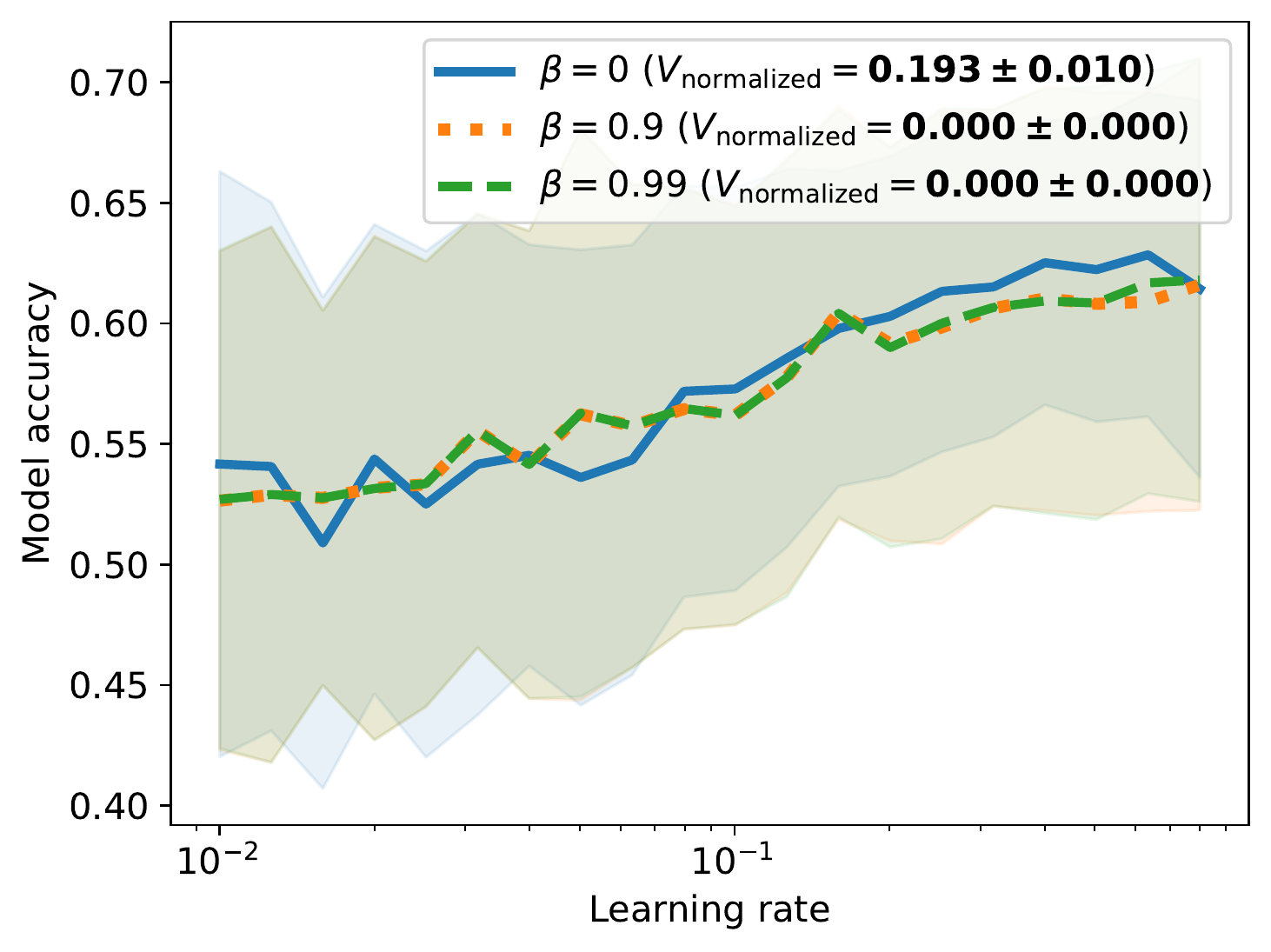}
      \caption{TCGA}
      \label{fig:rel_defense_acc_tcga}
   \end{subfigure}
   \begin{subfigure}[b]{0.4\textwidth}
      \centering
      \includegraphics[width=\textwidth]{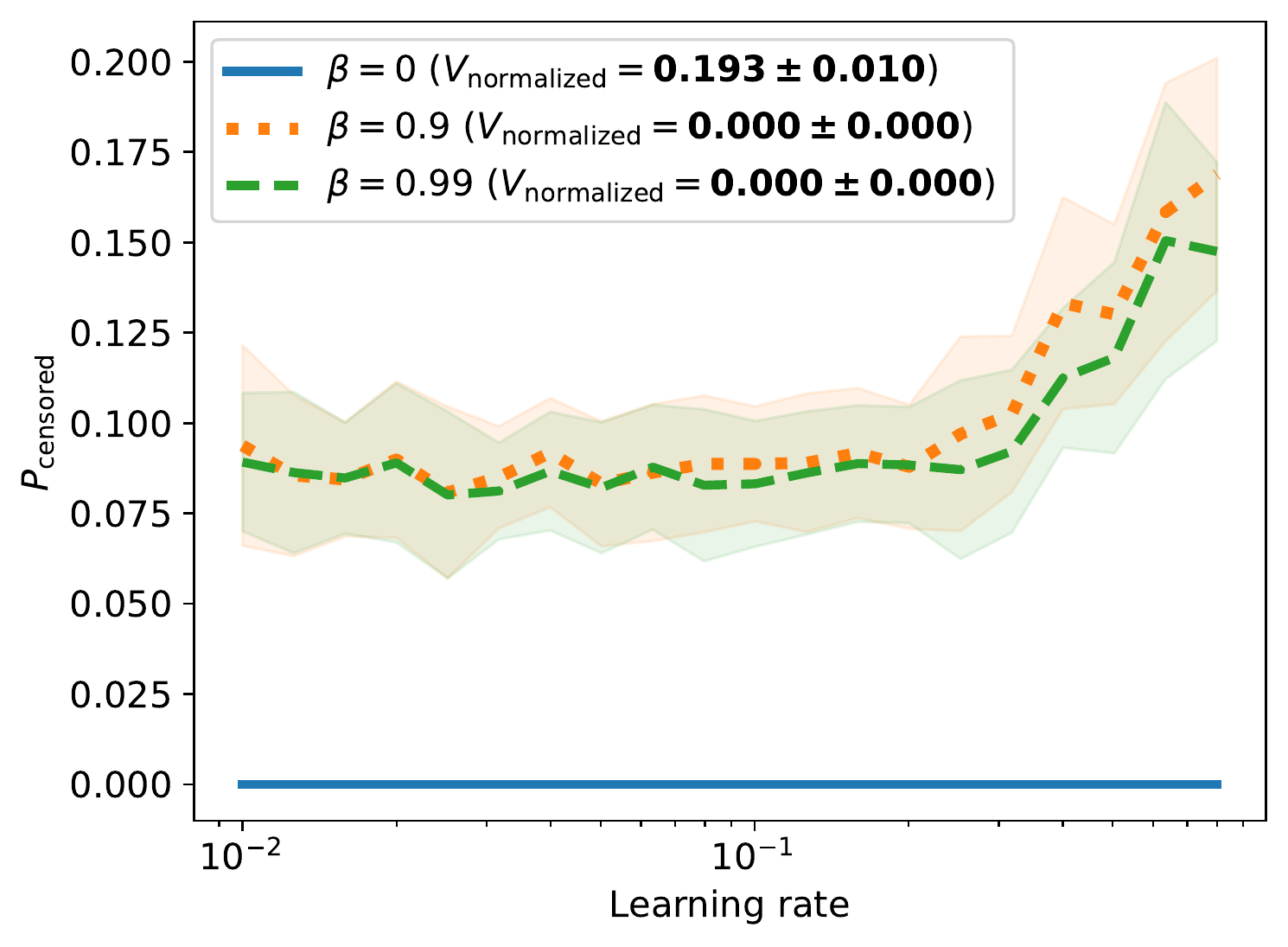}
      \caption{TCGA}
      \label{fig:rel_defense_prun_tcga}
   \end{subfigure}
      
   \caption{Effects of the  $\beta$-defense on DNA and TCGA (presented in~\Cref{sec:def_rel}) on the model accuracy and the number of neurons frozen. $P_\mathrm{censored}$ corresponds to the proportion of neurons censored compared to the total number of neurons.$\beta=0$ means that no defense is applied, neither $\beta$-defense nor $q$-defense.}
         \label{fig:rel_defense_acc_and_prun_all_datasets_2}
\end{figure}

\begin{table}[h]
   \centering\resizebox{\columnwidth}{!}{
   \begin{tabular}{lcccccc}
   \toprule
     Dataset  & $\rrecov \downarrow $   & $\rmatched \downarrow $   & $\rcomponent \downarrow $   & $\vweighed \downarrow $   & $P_\mathrm{censored}$   & Model Acc. $\uparrow$ \\
   \hline 
   CIFAR10 ($\beta=0.0$)       & $0.585 \pm 0.015$        & $0.514 \pm 0.021$      & $0.511 \pm 0.020$        & $0.490 \pm 0.019$         & $0.000 \pm 0.000$       & $0.261 \pm 0.039$ \\
   CIFAR10 ($\beta=0.9$)       & $\boldblue{0.000 \pm 0.000}$        & $\boldblue{0.000 \pm 0.000}$      & $\boldblue{0.000 \pm 0.000}$        & $\boldblue{0.000 \pm 0.000}$         & $0.094 \pm 0.013$       & $0.270 \pm 0.034$ \\
   CIFAR10 ($\beta=0.99$)      & $\boldblue{0.000 \pm 0.000}$        & $\boldblue{0.000 \pm 0.000}$      & $\boldblue{0.000 \pm 0.000}$        & $\boldblue{0.000 \pm 0.000}$         & $0.087 \pm 0.014$       & $\boldsymbol{0.271 \pm 0.033}$ \\

   DNA ($\beta=0.0$)           & $0.516 \pm 0.080$        & $0.031 \pm 0.018$      & $0.023 \pm 0.003$        & $0.233 \pm 0.032$         & $0.000 \pm 0.000$       & $0.929 \pm 0.017$ \\
   DNA ($\beta=0.9$)           & $\boldblue{0.000 \pm 0.000}$        & $\boldblue{0.000 \pm 0.000}$      & $\boldblue{0.000 \pm 0.000}$        & $\boldblue{0.000 \pm 0.000}$         & $0.020 \pm 0.007$       & $0.931 \pm 0.019$ \\
   DNA ($\beta=0.99$)          & $\boldblue{0.000 \pm 0.000}$        & $\boldblue{0.000 \pm 0.000}$      & $\boldblue{0.000 \pm 0.000}$        & $\boldblue{0.000 \pm 0.000}$         & $0.008 \pm 0.003$       & $\boldsymbol{0.932 \pm 0.021}$ \\

   FashionMNIST ($\beta=0.0$)  & $0.476 \pm 0.027$        & $0.285 \pm 0.022$      & $0.230 \pm 0.015$        & $0.284 \pm 0.015$         & $0.000 \pm 0.000$       & $\boldsymbol{0.732 \pm 0.053}$ \\
   FashionMNIST ($\beta=0.9$)  & $\boldblue{0.000 \pm 0.000}$        & $\boldblue{0.000 \pm 0.000}$      & $\boldblue{0.000 \pm 0.000}$        & $\boldblue{0.000 \pm 0.000}$         & $0.053 \pm 0.006$       & $0.725 \pm 0.042$ \\
   FashionMNIST ($\beta=0.99$) & $\boldblue{0.000 \pm 0.000}$        & $\boldblue{0.000 \pm 0.000}$      & $\boldblue{0.000 \pm 0.000}$        & $\boldblue{0.000 \pm 0.000}$         & $0.044 \pm 0.005$       & $0.727 \pm 0.043$ \\
   
   TCGA ($\beta=0.0$)          & $0.279 \pm 0.016$        & $0.199 \pm 0.014$      & $0.176 \pm 0.013$        & $0.193 \pm 0.010$         & $0.000 \pm 0.000$       & $\boldsymbol{0.660 \pm 0.065}$ \\
   TCGA ($\beta=0.9$)          & $\boldblue{0.000 \pm 0.000}$        & $\boldblue{0.000 \pm 0.000}$      & $\boldblue{0.000 \pm 0.000}$        & $\boldblue{0.000 \pm 0.000}$         & $0.169 \pm 0.032$       & $0.655 \pm 0.070$ \\
   TCGA ($\beta=0.99$)         & $\boldblue{0.000 \pm 0.000}$        & $\boldblue{0.000 \pm 0.000}$      & $\boldblue{0.000 \pm 0.000}$        & $\boldblue{0.000 \pm 0.000}$         & $0.147 \pm 0.025$       & $0.658 \pm 0.069$ \\
   \bottomrule
   \end{tabular}}
   \caption{Results on the relative defense presenting in~\Cref{sec:def_rel}. Influence of the relative defense on both the attack efficiency and the training performance. 
   The attack is performed on a simulation of a grid search, with 20 training with different learning rates. 
   The model accuracy reported is the best one for all the learning rates tested. This setting highlights 
   the defense success to prevent the attack for a range of learning rates, with no-significant loss of model accuracy. This table highlights that the relative defense ($\beta = 0.99$) is censoring less neurons that the absolute defense with $q=4$ for the same efficiency at defending from the attack ($\rrecov =0$).}
   \label{tab:def_rel_full}
   \end{table}

\subsection{Remark on preserving symmetry with $n_\mathrm{updates} =1$.}%
\label{app:def_n_update_1}
The second part of the attack is relevant only if $n_\mathrm{updates} >1$, which is the setting this paper focus on. 
It could be seen as a possible way of defending against \att. However enforcing $\nupdates =1$ by either (i) considering one batch in each round or (ii)
accumulating gradient at the initial point $\theta_{t-1}$, boils down to distributed SGD. This algorithm is not suitable in all FL settings as it comes with high communication costs~\cite{mcmahan2017communication}, and is more sensitive to gradient attacks, as studied by other work~\cite{geiping2020inverting,huang2021evaluating, hatamizadeh2022GradientInversionAttacks,xu2022agic}. As a result we do not explore this further.

\section{Hyper-parameters used in numerical experiments}
\Cref{tab:hp} lists the hyper-parameters used in the different numerical experiments shown in this paper.
\label{app:hp_results}
\begin{table}[h]
\centering
\begin{tabular}{lccccccccc}
\toprule
  Figure & Dataset & \# centers & \#$\mathcal{D}_k$ &   \# batch & \# hid. neur.& $n_\mathrm{updates}$ &  $t_\mathrm{max}$  &   \# trainings & lr\\
\hline 
\Cref{fig:main_result}&CIFAR10 &5&100&8&1000&5&20&20& $0.1$  \\
\Cref{fig:main_result}&FMNIST&5&100&8&1000&5&20&20& $0.5$  \\
\Cref{fig:main_result}&DNA&5&100&8&1000&5&20&20& $1.0$  \\
\Cref{fig:main_result}&TCGA&5&100&8&1000&5&20&20& $0.8$  \\
\Cref{fig:dirichlet,fig:dirichlet_extended}&FMNIST&5&100&8&1000&5&20&20& $0.5$  \\
\Cref{fig:effect_bs}&FMNIST&5&100&-&1000&5&20&20& $0.5$  \\
\Cref{fig:effect_neurons}&FMNIST&5&100&8&-&5&20&20& 0.5  \\
\Cref{fig:effect_lr}&FMNIST&5&100&8&1000&5&20&20& -  \\
\Cref{fig:effect_rounds}&FMNIST&5&100&8&1000&5&-&20& $0.5$  \\
\Cref{fig:effect_trainings}&FMNIST&5&100&8&1000&5&20&-& $0.5$  \\
\Cref{fig:effect_updates}&FMNIST&5&100&8&1000&-&20&20& $0.5$  \\
\Cref{fig:effect_num_datasamples}&FMNIST&5&-&8&1000&5&$0.2\#\mathcal{D}_k$&20&$0.5$  \\
\Cref{fig:effect_centers}&FMNIST&-&100&8&1000&5&20&20& $0.5$  \\
\Cref{fig:defense_acc_fmnist,fig:defense_prun_fmnist}&FMNIST&5&100&8&1000&5&20&20& - \\
\Cref{fig:rel_defense_acc_fmnist,fig:rel_defense_prun_fmnist}&FMNIST&5&100&8&1000&5&20&20& - \\
\Cref{fig:defense_acc_cifar10,fig:defense_prun_cifar10}&CIFAR10&5&100&8&1000&5&20&20& - \\
\Cref{fig:rel_defense_acc_cifar10,fig:rel_defense_prun_cifar10}&CIFAR10&5&100&8&1000&5&20&20& - \\
\Cref{fig:defense_acc_dna,fig:defense_prun_dna}&DNA&5&100&8&1000&5&20&20& - \\
\Cref{fig:rel_defense_acc_dna,fig:rel_defense_prun_dna}&DNA&5&100&8&1000&5&20&20& - \\
\Cref{fig:defense_acc_tcga,fig:defense_prun_tcga}&TCGA&5&100&8&1000&5&20&20& - \\
\Cref{fig:rel_defense_acc_tcga,fig:rel_defense_prun_tcga}&TCGA&5&100&8&1000&5&20&20& - \\

\bottomrule
\end{tabular}
\caption{Hyper-parameters used in the numerical experiments.}
\label{tab:hp}
\end{table}
\end{document}